\documentclass[10pt,twocolumn,letterpaper]{article}
\usepackage[pagenumbers]{cvpr}
\usepackage{booktabs} % For formal tables
\usepackage{pifont}
\usepackage{float}
\usepackage{tikz}
\usepackage{lipsum} % For formal tables
\usepackage[ruled]{algorithm2e} % For algorithms
\usepackage{makecell}
\usepackage{bm}
\usepackage{dsfont}
\usepackage{graphicx}
\usepackage{ulem}
\usepackage{mathrsfs} % in the preamble
\usepackage{multirow}

\newcommand{\argmin}{\operatornamewithlimits{argmin}}

\usepackage{colortbl} % Use colortbl specifically for table colors
\usepackage{soul} % Load soul for text highlighting

\newcommand{\supmatCOLOR}{black}

\newcommand{\supmat}{\textcolor{\supmatCOLOR}{{Sup.~Mat.}}\xspace}
\newcommand{\supmatlong}{\textcolor{\supmatCOLOR}{{Supplementary Material}}\xspace}
\newcommand{\supvid}{\textcolor{\supmatCOLOR}{{supplementary video}}\xspace}

\definecolor{first}{RGB}{255,178,178}
\definecolor{second}{RGB}{255,217,178}
\definecolor{third}{RGB}{255,255,178}

\newcommand{\methodname}{BimArt}

\newcommand{\cmark}{\textcolor{teal}{\ding{51}}}%
\newcommand{\xmark}{\textcolor{red}{\ding{55}}}%

\newcommand{\contact}{\mathbf{C}}
\newcommand{\contactseq}{\contact}
\newcommand{\numframes}{N}

\newcommand{\frameindex}{i}
\newcommand{\partindex}{p}
\newcommand{\handindex}{\rho}
\newcommand{\diffusionindex}{t}

\newcommand{\objectframe}{o}
\newcommand{\worldframe}{w}
\newcommand{\objecttraj}{\xi}
\newcommand{\objecttrajseq}{\xi}
\newcommand{\globalstate}{\mathbf{g}}
\newcommand{\globalstateseq}{\mathbf{G}}
\newcommand{\articulation}{\mathbf{a}}
\newcommand{\basis}{\mathbf{B}}
\newcommand{\basispoint}{\mathbf{b}}
\newcommand{\objectscale}{s_{\mathrm{o}}}

\newcommand{\objectvertex}{\mathbf{v}}
\newcommand{\objectvertices}{\mathbf{V}}
\newcommand{\objectverticesseq}{\mathbf{\objectvertices}}
\newcommand{\objectverticesOpen}{\mathbf{V}_\text{ao}}
\newcommand{\bpsprojecpoints}{\tilde{\mathbf{V}}}
\newcommand{\bpsprojecpointsseq}{\mathbf{\bpsprojecpoints}}

\newcommand{\handmotion}{\mathbf{H}}
\newcommand{\handmotionseq}{\handmotion}
\newcommand{\dirvec}{\mathbf{D}}
\newcommand{\dirvecseq}{\mathbf{\dirvec}}
\newcommand{\handvertices}{\mathbf{\Xi}}
\newcommand{\singlehandvertices}{\handvertices^{\handindex}}

\newcommand{\predhandverticesseq}{\hat{\handvertices}}
\newcommand{\handvertex}{\mathbf{h}}

\newcommand{\objectmotion}{\mathbf{O}}
\newcommand{\objectmotionseq}{\objectmotion}
\newcommand{\numhandpoints}{\mathtt{J}}
\newcommand{\numbasispoints}{\mathtt{K}}

\newcommand{\objecttoworld} {\mathbf{M}}
\newcommand{\manoparameters}{\mathbf{\Theta}}
\newcommand{\manoparametersseq}{\mathbf{\Theta}}
\newcommand{\manopose}{\boldsymbol{\theta}}
\newcommand{\manoshape}{\boldsymbol{\beta}}

\newcommand{\loss}{l}
\newcommand{\motiongen}{\mathcal{M}}
\newcommand{\latent}{\mathbf{Z}}
\newcommand{\latentobject}{\latent_o}
\newcommand{\latentcontact}{\latent_c}

\newcommand{\latentcombined}{\latent}
\newcommand{\encoder}{\mathcal{E}}
\newcommand{\contactobjencoder}{\encoder_o}
\newcommand{\objectencoder}{\encoder_{\alpha}}
\newcommand{\contactencoder}{\encoder_c}
\newcommand{\latentdim}{L}
\newcommand{\latentobjectdim}{\latentdim_o}
\newcommand{\latentcontactdim}{\latentdim_c}

\newcommand{\cfgdropoutprob}{p_f}
\newcommand{\guidancescale}{\lambda}
\newcommand{\cfgscale}{\guidancescale_f}
\newcommand{\cgscale}{\guidancescale_c}
\newcommand{\ppweight}{w}
\newcommand{\handkpdirvec}{\mathbf{X}}

\newcommand{\nulltoken}{\boldsymbol{\emptyset}}
\newcommand{\handprojpointsseq}{\mathbf{P}}
\newcommand{\handprojpoint}{\mathbf{p}}
\newcommand{\manoforward}{f_{\text{MANO}}}

\newcommand{\cams}{CAMS}
\newcommand{\camsx}{CAMS-X} % cross-category cams
\newcommand{\camsb}{CAMS-B} % cams in the bimanual setting
\newcommand{\mdmb}{MDM-B}
\newcommand{\omomob}{OMOMO-B}
\newcommand{\dmargin}{d_{\text{margin}}}

\usepackage{float}

\usepackage{xcolor}
\definecolor{forestgreen}{rgb}{0.13, 0.55, 0.13} % Soft green for R1
\definecolor{steelblue}{rgb}{0.27, 0.51, 0.71}   % Calm blue for R2
\definecolor{darkorange}{rgb}{1.0, 0.55, 0.0}    % Warm orange for R3

\definecolor{cvprblue}{rgb}{0.21,0.49,0.74}
\usepackage[pagebackref,breaklinks,colorlinks,allcolors=cvprblue]{hyperref}

\title{BimArt: A Unified Approach for the Synthesis of \\ 3D Bimanual Interaction with Articulated Objects}
\author{
Wanyue Zhang\textsuperscript{1,2}\quad
Rishabh Dabral\textsuperscript{1,2}\quad
Vladislav Golyanik\textsuperscript{1}\quad
Vasileios Choutas\textsuperscript{3}~~~\\
Eduardo Alvarado\textsuperscript{1}~~~~
Thabo Beeler\textsuperscript{3}~~~~~~~
Marc Habermann\textsuperscript{1,2}~~~
Christian Theobalt\textsuperscript{1,2}
\smallskip\\
\textsuperscript{1}MPI for Informatics, SIC\quad\quad
\textsuperscript{2}VIA Center\quad\quad
\textsuperscript{3}Google 
}

\begin{document}
\twocolumn[{
\renewcommand\twocolumn[1][]{#1}
\maketitle
\begin{center}
    \captionsetup{type=figure}
    \vspace{-5mm}
    \includegraphics[width=1.0\textwidth]{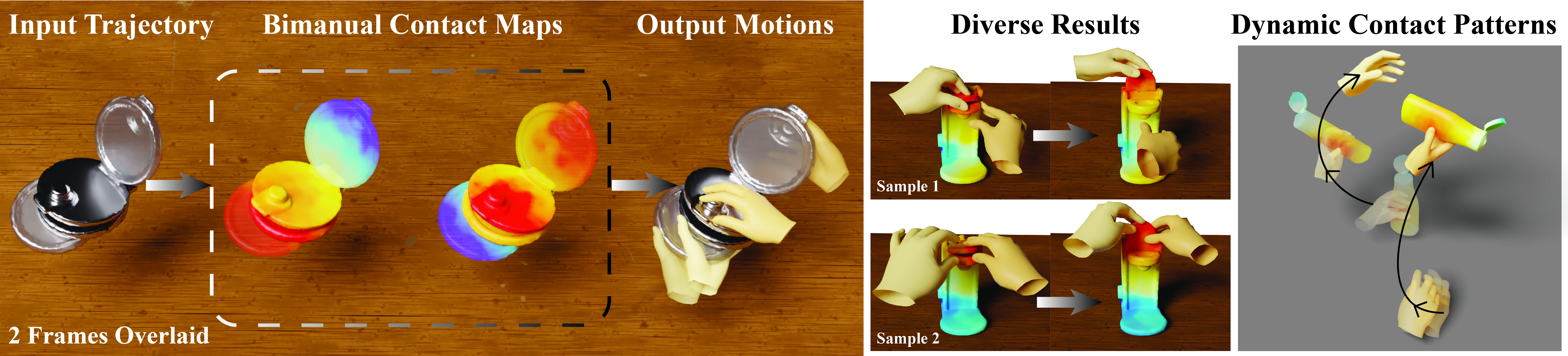}
	\caption
	{
    Given the mesh of an articulated object and its 7 DoF trajectories with 6D global states and 1D articulation, \methodname{} generates diverse and plausible hand motions that justify the object's trajectory. 
        Distance-based contact maps act as intermediate features for hand-object interaction, enabling our method to generate diverse and realistic bimanual motions. 
	}
    \label{fig:teaser}
\end{center}
}]
%
%%%%%%%%%%%%%%%%%%%%
%
\begin{abstract}
We present \methodname{}, a novel generative approach for synthesizing 3D bimanual hand interactions with articulated objects. Unlike prior works, we do not rely on a reference grasp, a coarse hand trajectory, or separate modes for grasping and articulating. To achieve this, we first generate distance-based contact maps conditioned on the object trajectory with an articulation-aware feature representation, revealing rich bimanual patterns for manipulation. The learned contact prior is then used to guide our hand motion generator, producing diverse and realistic bimanual motions for object movement and articulation. Our work offers key insights into feature representation and contact prior for articulated objects, demonstrating their effectiveness in taming the complex, high-dimensional space of bimanual hand-object interactions. Through comprehensive quantitative experiments, we demonstrate a clear step towards simplified and high-quality hand-object animations that surpass the state of the art in motion quality and diversity. Project page: \url{https://vcai.mpi-inf.mpg.de/projects/bimart/}.
\end{abstract}
%
%%%%%%%%%%%%%%%%%%%%
%
%
%%%%%%%%%%%%%%%%%%%%
%
\vspace{-5mm}
\section{Introduction} \label{sec:1-introduction}
%
% Motivation
%
Humans engage with articulated objects in countless ways throughout the day,  whether it is twisting the cap of a water bottle, tilting a laptop screen for better viewing, or deftly slicing through paper with a pair of scissors.
Although these interactions seem effortless for humans, they are challenging to generate computationally due to the highly complex and high-dimensional space of bimanual hand animations that not only rigidly move an object, but also generate meaningful object articulations.
From a 3D modeling perspective, these motions require a deeper understanding of the individual object parts, their interaction affordances, and their geometry.
%
%%%%%%
%
% Prior work
%
\par
Despite substantial progress in 3D character animation research~\cite{starke2019neural, zhang2022couch, li2023object, hassan2021stochastic} driven by deep learning and generative models, recent works either focus on synthesizing whole-body without considering hands~\cite{xu2023interdiff, starke2019neural}, or generate hand-object interaction assuming objects are \textit{rigid}~\cite{ghosh2022imos, christen2022dgrasp, zhang2024graspxl}. 
Very few studies address 3D bimanual interactions with \textit{articulated} objects.
Methods that are designed for articulated objects either work in a category-specific manner~\cite{Zheng_2023_CVPR_cams} for unimanual motions, cannot simultaneously perform articulation and object root translation and rotation~\cite{zhang2024artigrasp}, or rely on noisy hand-object interaction sequences as input and refine hand motions afterwards~\cite{liu2024geneoh}.
Some works~\cite{zhang2024artigrasp, ghosh2022imos, Zheng_2023_CVPR_cams} also assume the initial or goal grasp to be known, which can be a restrictive assumption for non-expert users.
%
%%%%%%
%
% Method (high level)
%
\par
In contrast to prior works, \methodname{} operates with relaxed assumptions: it does not assume a known reference grasp, is not trained in an object-specific manner, does not require a coarse hand trajectory, and can perform object articulation simultaneously with the object's root rotation and translation (see a conceptual comparison in Tab.~\ref{tab:concecptual_comparison} in the \supmatlong).
Given object trajectories, which involve global translation, rotation, and articulation, \textbf{\methodname} generates diverse and realistic bimanual motions for grasping and articulating the object (see Fig.~\ref{fig:teaser}).
We propose a three-stage approach:
Our Bimanual Contact Generation model first generates contact maps, capturing dynamic interactions between the hand and the object over time.
Next, the generated contact maps and the object geometry are used as conditionings to synthesize hand animations using our generative Bimanual Motion Model. 
Finally, we refine the generated animations with contact guidance, followed by explicit optimization to remove artifacts like penetration or missing hand-object contact.
%%%%%%
%
% Contact maps
%
\par
To be category-agnostic and to accommodate a wide array of geometries, we propose a novel articulation-aware representation based on basis point sets (BPS)~\cite{prokudin2019efficient}, originally defined as a collection of vectors from a fixed set of points in space to the nearest vertices of the object.
Our representation involves normalizing the object’s scale and then computing the distance vectors from the BPS to each articulated part independently. 
This part-based representation treats each component of an object equally, ensuring that different surface areas have similar spatial encoding resolution.
Given the above object encoding, our contact generation network predicts distance-based bimanual contact maps, which serve as an intermediate generation target, removing the need for a reference grasp.
Our key insight is that frame-wise contact maps embedded on the object capture diverse grasping patterns and offer more nuanced and detailed information compared to sparser or stage-wise contact points~\cite{li2023object, peng2023hoi, Zheng_2023_CVPR_cams} for bimanual interaction synthesis. 
We evaluate \methodname{} on ARCTIC~\cite{fan2023arctic} and HOI4D~\cite{Liu_2022_CVPR} datasets and achieve state-of-the-art performance in terms of interaction plausibility and diversity. 
%
%%%%%%
%
% Summary
%
\par
To summarize, our contributions are as follows:
%
%%%%
\begin{itemize}
    \item BimArt, a new approach for bimanual hand motion synthesis for interaction with articulated objects;
    \item A canonicalized and part-aware object feature representation, which is able to encode diverse and articulated objects in a unified representation well suited for object-aware hand animation synthesis;
    \item A generative model for bimanual contact maps that serve as an interaction prior for our hand motion synthesizer. 
\end{itemize} 
%%%%
%
%%%%%%%%%%%%%%%%%%%%
%
%
%%%%%%%%%%%%%%%%%%%%
%
\section{Related Work} \label{sec:2-related-work}
%
%%%%%%%%%%%%%%%%%%%%
%
\par \textbf{3D Human Motion Synthesis.}
3D human motion synthesis is an active and long-standing research field \cite{stylemachines2000, bilinear_spatiotemporal_2012, gpmotion2008, lehrmann2014efficient, Fragkiadaki_2015_ICCV, Martinez_2017_CVPR, holden_phase_2017,petrovich2021action, deepphase2022, petrovich24stmc, braun2023physically, zhang2024roam}. 
Over the last years, neural-network-based methods have dominated it, aided by the availability of large-scale datasets of body-only \cite{AMASS:2019,ionescu2013human3}, hands-only \cite{moon2020interhand2,Hampali_2020_CVPR,Chao_2021_CVPR}, or whole-body motion \cite{GRAB:2020,fan2023arctic}. 
Diffusion models have manifested their potential to generate diverse and high-quality motions 
using conditioning signals such as text, audio, scene context, or the movements of other people~\cite{dabral2022mofusion, mughal2024convofusion, tevet2023human, yi2024tesmo, petrovich24stmc, ghosh2023remos, karunratanakul2023dno, xu2023interdiff, li2023object, li2023controllable, Menapace2024ToG}. 
All these methods synthesize full-body, while hand-object interaction requires more fine-grained consideration of joint movement and alignment with object geometry. 
%
%%%%%%%%%%%%%%%%%%%%
%
\par \textbf{Hand-Object Interaction.}
Similar to 3D human motion synthesis, the introduction of hand-object interaction datasets ~\cite{GRAB:2020, Brahmbhatt_2019_CVPR, Liu_2022_CVPR,fan2023arctic,liu2024taco} has led to rapid developments in 3D hand and object pose reconstruction~\cite{ye2022hand, ye2023vhoi, zhang2024ddf, zhu2023contactart, HMP}, static grasp synthesis~\cite{lee2024interhandgen, ye2024g, Turpin_graspd, jiang2021hand, tendulkar2023flex, karunratanakul2020grasping, liu2023contactgen}, hand-object interaction (HOI) motion denoising~\cite{zhou2022toch, liu2024geneoh, hao2024hand}, and dexterous object manipulation in robotics~\cite{lee2024dextouch, wang2022dexgraspnet, wan2023unidexgrasp++, chi2023diffusionpolicy, yuan2023robot, huang2023dynamic, wang2024cyberdemo, turpin2023fastgraspd}. 
However, except for the concurrent work ManiDext~\cite{zhang2024manidext}, existing methods~\cite{taheri2021goal, Zheng_2023_CVPR_cams, zhou2024gears, zhang2021manipnet, ghosh2022imos, cha2024text2hoi, MACS2024} either generate single hand motions, do not work with articulated objects~\cite{taheri2021goal, zhou2024gears, zhang2021manipnet, ghosh2022imos, MACS2024}, or rely on different input assumptions such as hand trajectories~\cite{zhang2021manipnet} or textual task descriptions~\cite{cha2024text2hoi, christen2024diffh2o, peng2023hoi}. 
Among works that show applicability in articulated objects, text conditioning~\cite{cha2024text2hoi} lacks fine-grained control over object paths that is often essential in artistic creation.
ArtiGrasp~\cite{zhang2024artigrasp} requires a reference pose and cannot handle grasping and articulation simultaneously.
CAMS~\cite{Zheng_2023_CVPR_cams} relies on the initial grasp as input and trains a separate model per category. 
In contrast, we train a unified model for all categories and do not rely on reference poses.
%
%
%%%%%%%%%%%%%%%%%%%%
%
\par \textbf{HOI Feature Representation.}
Existing HOI feature representations~\cite{cha2024text2hoi, zhang2021manipnet, Zheng_2023_CVPR_cams, zhou2024gears,ye2024g} are either not suitable for motion synthesis or fail to emphasize the articulated structure of objects, which is our focus. 
ManipNet~\cite{zhang2021manipnet} utilizes a coarse voxel-based representation to capture the object's global geometry for rigid objects.
\cams~\cite{Zheng_2023_CVPR_cams}'s stage-wise contact target design struggles to capture the rich bimanual interaction patterns.
Works focusing on motion denoising~\cite{zhou2022toch, liu2024geneoh} compute detailed spatiotemporal features, such as motion velocities and contact correspondence.
However, generating these features from scratch without assuming an initial motion is challenging and may overconstrain the synthesis model, leading to lower diversity.
In contrast to these previous works, we propose a part-based object representation specifically designed for articulated objects, ensuring that objects with unbalanced part sizes are not disadvantaged. 
Additionally, our hand representation encodes both surface positions and distances to the object, enhancing interaction plausibility.
%
%%%%%%%%%%%%%%%%%%%%
%
% Double column figure.
 \begin{figure*}[!t]
 \centering
	\includegraphics[width=\linewidth]{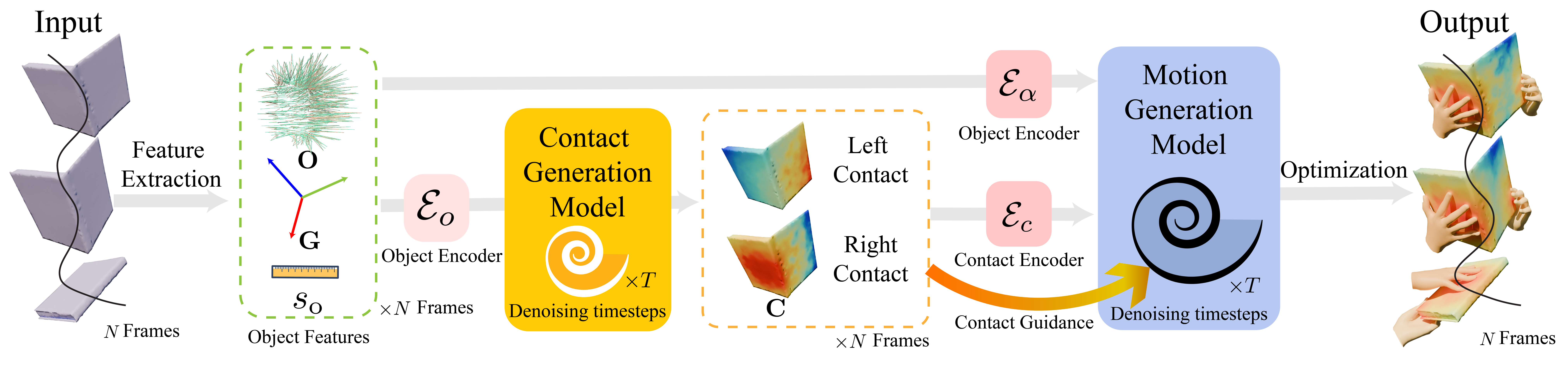}
        \caption
	{
        \textbf{Overview of the proposed approach.} 
         \methodname{} takes $\numframes$ frames of object trajectories as input and generates $\numframes$ frames of 3D bimanual interactions. The object features (articulation-aware BPS features 
        $\objectmotionseq$, 6D global states $\globalstateseq$, and the object scale $\objectscale$) are passed into both the object encoder $\contactobjencoder$ (MLP) in the contact generation model and $\objectencoder$ (MLP) in the motion generation model. 
         Additionally, the motion generation model's contact encoder $\contactencoder$ takes
        $\contactseq$, the bimanual contact map produced by the contact generation model, as conditioning input.
        The contact model and motion model are both denoising diffusion models, and the spiral denotes the denoising process.
        $\contactseq$ is further used as guidance at each diffusion timestep to align hand motions with the generated contact maps. 
        Finally, we use optimization to correct contact and penetration artifacts and obtain 3D bimanual meshes.
        }
	\label{fig:overview}
\end{figure*}

\section{Method}
Our goal is to generate realistic, diverse, and contact-aware
3D bimanual motion from a sequence of articulated object states.
We consider two-part articulated objects with a total of seven degrees of freedom: six degrees for the root's orientation and translation, and one degree for the rotational joint.
The input to our method is the articulated object trajectory, $\objecttrajseq = \{\objecttraj_i\}_{i=1}^N, \objecttraj_i = [\globalstate_i | \articulation_i]$, where $\globalstate_i \in \mathbb{R}^6$ denotes the object's orientation and global translation, and $\articulation_i \in \mathbb{R}$ represents the articulation angle between the two parts of the object. 
Given $\objecttrajseq$, \methodname{} generates a corresponding, $\numframes$-frame bimanual motion  $\manoparametersseq = \{\manoparameters_\frameindex\}_{\frameindex=1}^{\numframes}$, where $\manoparameters_i \in \mathbb{R}^{61 \times 2}$ corresponds to MANO~\cite{MANO:SIGGRAPHASIA:2017} hand parameters for both hands.
\par
\cref{fig:overview} outlines our method.
We first introduce an articulation-aware canonicalized feature representation for the object (Sec.~\ref{sec:feature}).
By keeping the canonicalized object at the origin of the coordinate system, we provide a consistent frame of reference for the object as well as the hands.
Next, motivated by the observation that contact understanding facilitates more accurate finger placement, we decompose the task into contact map generation (Sec.~\ref{sec:contact}) and motion synthesis based on the generated contact map (Sec.~\ref{sec:motion}). 
Lastly, we use an optimization-based post-processing step to resolve physical artifacts such as penetration and inconsistent contact. (Sec.~\ref{sec:postprocess}). 
%
%
%%%%%%%%%%%%%%%%%%%%
%
\subsection{Hand and Object Representation} \label{sec:feature}
\par
\begin{figure}[!hbt]
\vspace{-1cm}
	\includegraphics[width=0.9\linewidth]{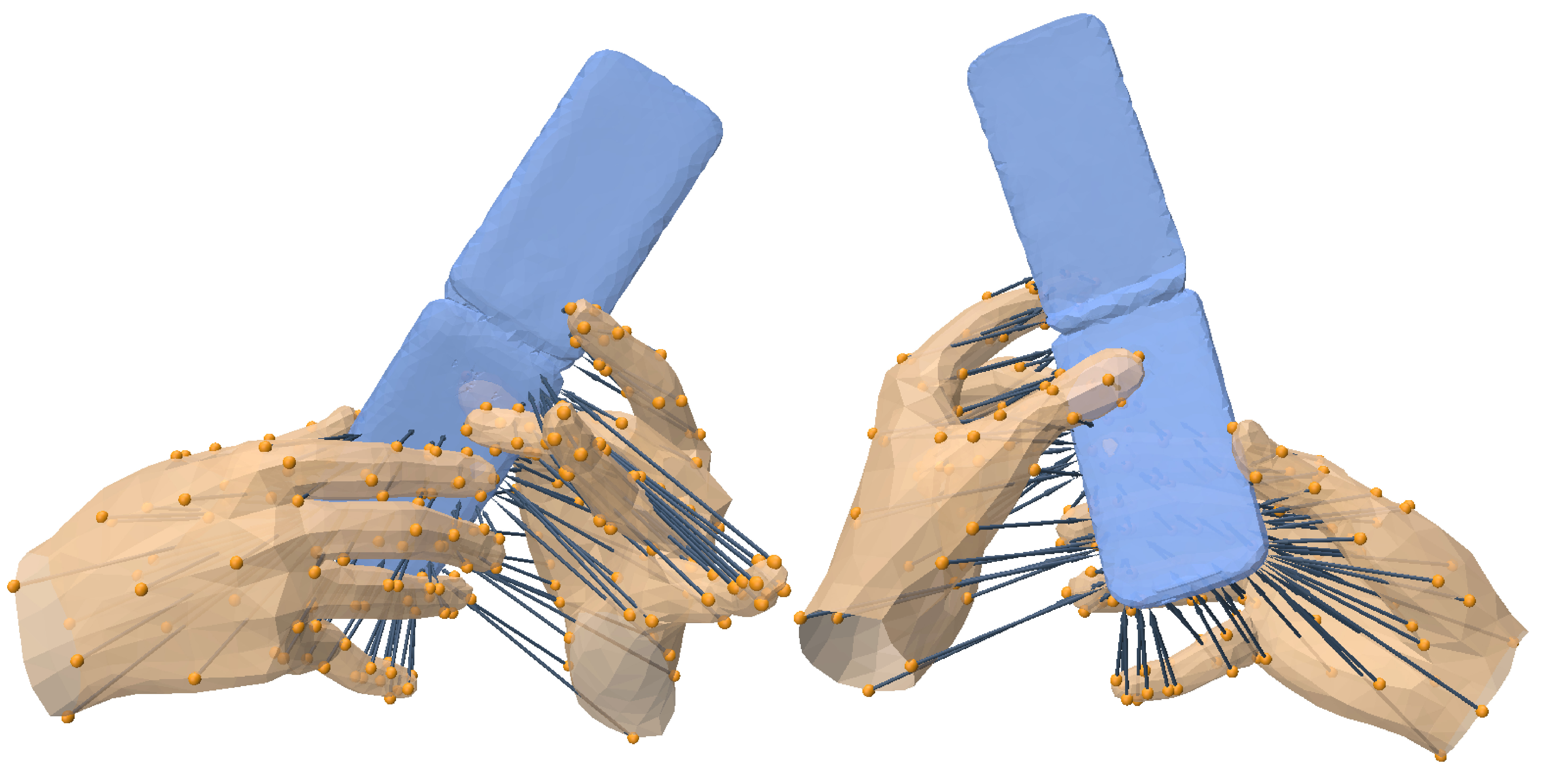}
	\caption{\textbf{Hand Representation:} We parameterize each frame of hand pose by using $\numhandpoints$ surface keypoints (in orange), sampled from the surface of the hand. In addition to position, we also use the direction vector (dark blue lines) from each keypoint to the nearest object surface as an additional feature.
	\label{fig:model_features}
 }
 \vspace{-5mm}
\end{figure}

 \textbf{Hand Representation.}
We encode hand motion in an object-centric way, with each hand at frame $\frameindex$ parameterized by both surface keypoint positions $\handmotion_\frameindex$ and direction vectors to the object $\dirvec_\frameindex$  as shown in ~\cref{fig:model_features}.
%
%%%%
%
%%%%
%
More specifically, 
$\handmotion_\frameindex \in \mathbb{R}^{\numhandpoints \times 3}$ is a sparse set of vertices sampled from the MANO surface vertices $\handvertices_\frameindex$.
$\dirvec_\frameindex \in \mathbb{R}^{\numhandpoints \times3}$ denotes the direction vectors originating from the hand keypoints $\handmotion_\frameindex$ to their nearest object vertices.
$\dirvec_\frameindex$ encodes both the direction and the magnitude.
Compared with the MANO skeletal joints, this representation is denser, making it easier to recover MANO parameters, $\manoparameters_\frameindex$.
In addition, the incorporation of $\dirvec_i$ aids the model in reasoning about contact. 
\par
To generalize to unseen object trajectories and disentangle
object motion due to articulation and global trajectory changes, we propose to encode the hand in the object's canonical coordinate frame, \ie the frame where the object's articulation axis is aligned with the negative $z$-axis.
Let $\objectvertices_\frameindex$ denote the object vertex positions at frame $i$, 
and $\objecttoworld$ be its canonical-to-world transformation matrix.
We transform the hand point cloud and object vertices from the world frame $\handmotion_\frameindex^\worldframe$ to the object's canonical frame $\handmotion_\frameindex^\objectframe$ such that
$
     \handmotion_\frameindex^{\objectframe} = \left(\objecttoworld\right)^{-1} 
    \handmotion_\frameindex^{\worldframe} $ and $\objectvertices_\frameindex^{\objectframe} = \left(\objecttoworld\right)^{-1} 
    \objectvertices_\frameindex
$.
In the rest of the paper, we omit $\objectframe$, since all hand motions are generated in the object's canonical frame. 
\par \textbf{Object Representation.}
Next, we define the object feature representation. Training a single model across multiple object types necessitates a feature representation that encodes geometric information consistently while remaining independent of the object topology.
We therefore represent the object trajectory using Basis Point Sets (BPS) \cite{prokudin2019efficient}.

The BPS representation requires defining a fixed set of basis points $\basis \in \mathbb{R}^{\numbasispoints \times 3}$ 
that are typically uniformly sampled from the unit sphere.
The BPS features are then computed as a set of vectors from $\basis$ to the nearest object vertices.
This formulation will lead to a suboptimal sampling strategy for objects with articulated parts at different scales.
In contrast to the original BPS formulation, we propose to use normalized part-based BPS features computed in the object-centric frame where the same basis point set is mapped separately to each articulated part. 
Let $\objectscale$ denote the object scale, computed by normalizing the maximum distance from the origin to the object vertices ($\objectvertices_{ao}$) with an open articulation angle in the canonical space:
\begin{equation}
\objectscale = \frac{1 - \dmargin}{\max_{\mathbf{\objectvertex} \in \objectverticesOpen} \|\mathbf{\objectvertex}\|}
\end{equation}
A margin $\dmargin$ is used to prevent the object point cloud from touching the boundary of the unit ball.
To provide a denser mapping from basis points $\basis$ to object vertices $\objectvertices$, the object is normalized to the unit sphere, using $s_o$.
The normalized, part-based BPS features are computed as:
\begin{equation}\label{eqn:part_bps}
    \objectmotion_\frameindex^\partindex = \left [
    \argmin_{\objectvertex \in \objectvertices_\frameindex^\partindex} 
    d(\frac{\objectvertex}{\objectscale} , \basispoint) - \basispoint
    \text{ , for } \basispoint \in \basis \right ]
\end{equation} 
\begin{equation}
    \objectmotionseq = \left [ \objectmotion_\frameindex^p \text{ , for } \frameindex \in \{1, 2, \ldots, \numframes\}, \partindex \in \{\text{top}, \text{bottom}\} \right ]
\end{equation}
In \cref{eqn:part_bps}, $d(\cdot)$ is the Euclidean distance between two points and $\partindex$ denotes the part index. 
Notably, we do not perform a part-based scale normalization, since hand motion $\handmotion$ is encoded in the original scale, and having separate object scales in canonical spaces will increase the difficulty for the model in reasoning about hand object distance and contact.
\par
\begin{figure}[!t]
	\includegraphics[width=\linewidth]{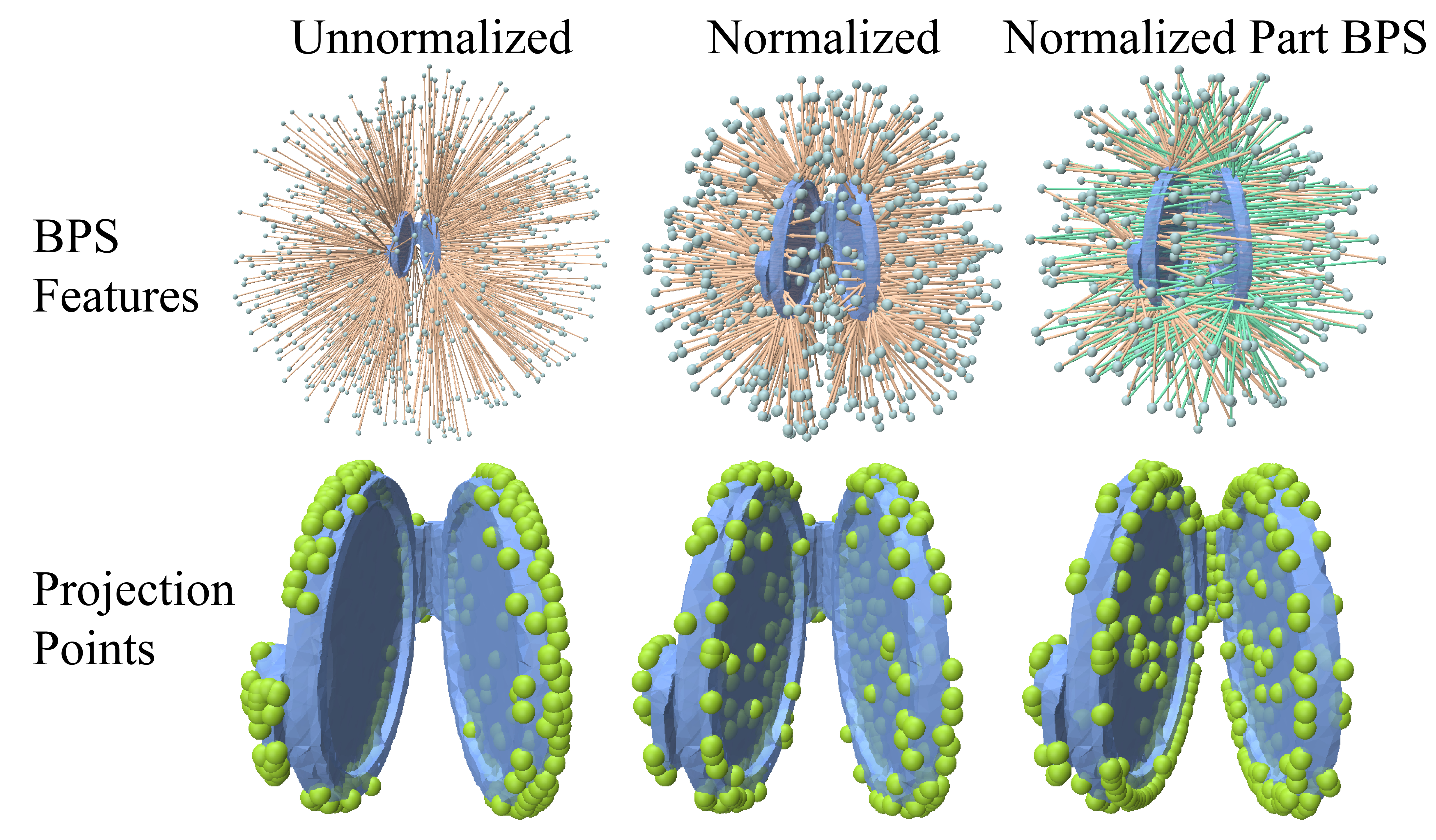}
	\caption{
        \textbf{Different BPS Sampling Strategies}. 
Top left: $\numbasispoints \times 2$ basis points sampled uniformly within a 0.5-meter radius for unnormalized objects. Top middle: $\numbasispoints \times 2$ BPS sampled uniformly in a unit ball for normalized objects. Top right: $\numbasispoints$ basis points sampled uniformly in a unit ball for normalized objects, with points mapped to each articulated part of the object, maintaining the same feature dimension. Bottom: Green points on the object represent the projections of the BPS feature vectors. The proposed Normalized Part BPS provides denser mapping on the object's inner surface layer.}
	\label{fig:bps_features}
\vspace{-5mm}
\end{figure}

Alternatively, one could sample the basis points in the original scale of the object without normalizing the object to a unit sphere or using a part-agnostic BPS mapping;
the comparison for different sampling strategies is shown in \cref{fig:bps_features}.
Our part-based BPS with scale normalization provides a denser mapping to the object, thus forming a more detailed descriptor of the object geometry.
It ensures that the objects with a small articulating part (e.g.,~the lid of a bottle) are not under-sampled against the larger base.
\par
Our BPS feature $\objectmotionseq$ is independent of the object's global trajectory, encoding only the object's shape and articulation states.
However, without encoding the object's global movement, the generated motion will be physically implausible since it is not aware of the gravity direction and cannot distinguish the object trajectories that require a supporting hand at the bottom. 
Therefore, we further include the global states $\globalstateseq = [\globalstate_\frameindex]_{\frameindex=1}^{\numframes}$ as a lower dimensional 6D vector per frame, which consists of relative translation to the first frame and the global rotation. 
Relative translation is used to avoid overfitting and increases robustness to unseen test trajectories. 
Overall, $\objectmotionseq$ and $\globalstateseq$ capture the detailed object geometry, articulation movement, and global movement. 
%
%
%%%%%%%%%%%%%%%%%%%%
%
\subsection{Bimanual Contact Generation Model}
\label{sec:contact}
Having defined our articulated object representation, we introduce our novel denoising diffusion probabilistic model for generating plausible bimanual contact maps.
Importantly, our contact model can be jointly trained on \textit{cross-category} articulated objects, thanks to our generalizable object representation.
%
%%%%%%
%
\par 
Given our object BPS features $\objectmotionseq$, global states $\globalstateseq$, and the object scale $\objectscale$, 
the contact model generates the corresponding sequence of contact maps for the left and right hand, \ie $\contactseq = [\contactseq^\handindex], \handindex \in \{\text{left}, \text{right}\}$.
\par
Our bimanual contact maps at frame $\frameindex$ are defined as the minimum distance from each object vertex $\objectvertex$ from any of the hand vertices $\singlehandvertices_\frameindex$ for each hand:
\begin{equation}\label{eqn:contact}
    \contact^\handindex_\frameindex = \left[
    \argmin_{\handvertex \in \singlehandvertices_\frameindex} d(\handvertex, \objectvertex) - \objectvertex \text{ , for } \objectvertex \in \bpsprojecpoints_\frameindex 
    \right], \handindex \in \{\text{left}, \text{right}\}
\end{equation}
where $\bpsprojecpoints_\frameindex = \objectmotion_\frameindex + \basis$, the closest object vertices from the basis points $\basis$.
We generate separate contact maps for the left and right hand to reduce ambiguity when using them as guidance in motion generation (see~\cref{sec:motion}). 
Note, that our contact map does not encode correspondence between which hand vertex should be in contact with the object vertex, as doing so would over-constrain the sampling process, thereby hindering motion diversity. 
\par 
For contact generation, we adopt a denoising diffusion probabilistic model~\cite{sohl2015deep, ho2021classifierfree} with a transformer-encoder architecture~\cite{tevet2023human, vaswani2017attention}, trained to directly predict clean samples $\contactseq$.
The model's conditioning inputs include our BPS features $\objectmotionseq$, global states $\globalstateseq$, and $\objectscale$, which are processed through an MLP encoder, $\contactobjencoder$. 
We predict a contact value per BPS feature in $\objectmotionseq$, facilitating cross-object predictions as the output dimension is fixed by the number of BPS points and remains independent of the object's mesh resolution as shown in \cref{eqn:contact}.
Next, we show how these generated contact maps are used to synthesize hand motions (Sec.~\ref{sec:motion}).
%
%%%%%%%%%%%%%%%%%%%%
%
\subsection{Bimanual Hand Motion Model}
\label{sec:motion}
Given the object features $\objectmotionseq$ and the contact maps $\contactseq$, our motion model generates $\numframes$ frames of hand motions, parameterized by $\handkpdirvec = [\handmotionseq | \dirvecseq]$ as illustrated in Fig.~\ref{fig:overview}.
%
%%%%%%
Specifically, the conditions include:
\begin{itemize}
    \item Object Conditioning: The BPS features $\objectmotionseq$, object global states $\globalstateseq$, and the object scale $\objectscale$
    are encoded using the object encoder $\objectencoder$ into a latent object embedding $\latentobject \in \mathbb{R}^{\numframes \times \latentobjectdim}$, where $L_o$ denotes the latent dimension. 
    \item Contact Conditioning: The contact maps $\contactseq$ are encoded using the contact encoder $\contactencoder$ into a latent contact feature embedding $\latentcontact \in \mathbb{R}^{\numframes \times \latentcontactdim}$.
\end{itemize}
$\objectencoder$ and $\contactencoder$ are MLPs and their respective outputs are concatenated to form $\latentcombined = [\latentobject | \latentcontact ] $.
\par
Similar to the contact model, we use another transformer encoder (denoted as $\motiongen$) as the diffusion denoiser that is also trained to predict the clean samples.
To learn smooth and diverse hand motions and counter the potential noise in the contact model's prediction, we train $\motiongen$ using classifier-free guidance by randomly replacing the contact features $\latentcontact$ with a learnable null token $\nulltoken$ with a probability $\cfgdropoutprob$. %
\par
We observe that $\latentcontact$ effectively guides $\motiongen$ to establish and maintain contact with the articulated object, dynamically adjusting to changing contact patterns while ensuring temporal consistency.
However, at a more fine-grained level, the generated motion is not free from physical artifacts such as fingers being stuck between the parts.
Therefore, we introduce a contact map discrepancy term during guidance, encouraging the noisy motion at each denoising timestep to more precisely align with the contact map output $\hat{\contactseq}$ from the contact model.
Namely, for each predicted clean hand $\hat{\handmotionseq}_{(\diffusionindex)}^\handindex$ at denoising timestep $\diffusionindex$, we compute a derived contact map $\tilde{\contactseq}_{(\diffusionindex)}^\handindex$ from $\hat{\handmotionseq}_{(\diffusionindex)}^\handindex$ to the nearest object vertex:
\begin{equation}
    \tilde{\contactseq}_{(\diffusionindex)}^\handindex = \left [
    \argmin_{\handvertex \in {\hat{\handmotionseq}}_{(\diffusionindex)}^\handindex} d(\handvertex, \objectvertex) - \objectvertex \text{ , for } \objectvertex \in \bpsprojecpointsseq
    \right], \handindex \in \{\text{left}, \text{right}\}
\end{equation}
In practice, we use a differentiable one-nearest-neighbor function for the above computation to ensure gradient propagation.
The contact map guidance can be written as
\begin{equation}
\tilde{\handkpdirvec}_{(\diffusionindex)}^\handindex = \hat{\handkpdirvec}_{(\diffusionindex)}^\handindex - \cgscale \nabla_{\hat{\handkpdirvec}_{\diffusionindex}^\handindex} \left\Vert \hat{\contactseq}^\handindex - \tilde{\contactseq}_{(\diffusionindex)}^\handindex \right\Vert,
\end{equation}
where $\cgscale$ is the guidance scale and $\nabla_{\hat{\handkpdirvec}_{\diffusionindex}^\handindex}$ denote the gradient with respect to the discrepancy term. 
Finally, we apply classifier-free guidance to combine the outputs with and without $\latentcontact$.
The predicted clean motion at timestep $\diffusionindex-1$ can be written as
\begin{equation}
\tilde{\handkpdirvec}_{(\diffusionindex-1)} = (1 + \cfgscale) \tilde{\handkpdirvec}_{(\diffusionindex)} -\motiongen(\hat{\handkpdirvec}^{(\diffusionindex)}, \diffusionindex, \latentobject \nulltoken).
\end{equation}
Since $\motiongen$ does not predict dense MANO surface vertices, we rely on an optimization-based MANO fitting described in the next section to obtain the final 3D bimanual motions. 

\subsection{Physically Plausible Hand Motion}\label{sec:refinement}
\label{sec:postprocess}
Our generated hand motions $\hat{\handmotionseq}$ only contain a subset of MANO surface vertices and, therefore, some hand surface areas may still experience minor penetration, momentary loss of contact, or slight jitter after denoising.
To address this, we introduce an optimization-based MANO fitting to further refine the predictions. 
\par
First, we estimate the MANO parameters $\manoparameters =\left[\manopose | \manoshape\right]$ for both hands,
where $\manopose \in \mathbb{R}^{\numframes \times 51 \times 2}$ and
$\manoshape \in \mathbb{R}^{10 \times 2}$,
from the predicted $\hat{\handmotionseq}$. $\manopose$
contains the root translation, rotation, and per-joint rotations of MANO, with all rotations represented as axis-angle vectors. 
We estimate $\manoparameters$ by minimizing the following loss:
\begin{equation}
    \loss_{\text{MANO}} = 
    \lVert \hat{\handmotionseq} -
    \manoforward(\manopose,\manoshape)
    \rVert,
\end{equation}
$\manoforward$ is the MANO forward pass operator to retrieve fitted hand keypoints based on optimized $\manopose,\manoshape$.
\par
Next, we refine the estimated MANO parameters $\manoparameters$ to reduce penetrations and temporal jitter and enforce
contact at the predicted points using three energy terms: 
\begin{equation}
        \loss_{\text{reg}} = 
          \ppweight_{\text{proj}} \loss_{\text{proj}}
        + \ppweight_{\text{pen}} \loss_{\text{pen}}
        + \ppweight_{\text{acc}} \loss_{\text{acc}}.
\end{equation}
Since our denoising outputs contain both $\hat{\handmotionseq}$ and $\hat{\dirvecseq}$, the projection loss $\loss_{\text{proj}}$ encourages the projection points of hand keypoints based on the direction vectors,  \ie. $\handprojpointsseq = \manoforward(\manopose,\manoshape) + \hat{\dirvecseq}$, to lie on the object surface, resolving the potential floating artifact.  
\begin{equation}
   \loss_{\text{proj}} = \sum_{
   \handprojpoint \in \handprojpointsseq
   }
   \min_{\objectvertex \in \objectverticesseq}
   \left\lVert \handprojpoint - \objectvertex \right\rVert.
\end{equation}
The dense predicted hand surface vertices after MANO fitting, denoted as $\predhandverticesseq$, are fed into the penetration loss~\cite{hasson19_obman}:
\begin{equation}
\loss_{\text{pen}} = \sum_{
\handvertex \in \mathrm{Int}(\predhandverticesseq)
}
\min_{\objectvertex \in \objectverticesseq}
\left\lVert \handvertex - \objectvertex \right\rVert.
\end{equation}
$\mathrm{Int}(\predhandverticesseq)$ refers to the set of hand vertices inside the object.
Finally, we penalize the acceleration of hand vertices $\predhandverticesseq$ :
\begin{equation}
   \loss_{\text{acc}} = 
     \sum_{\handvertex_\frameindex \in \predhandverticesseq} \left\lVert \handvertex_\frameindex  - 2 \cdot \handvertex _{\frameindex-1} + \handvertex_{\frameindex-2} \right\rVert,
\end{equation}
$w_{\text{proj}}$, $w_{\text{pen}}$ and $w_{\text{acc}}$ are hyperparameters. 
We demonstrate the effectiveness of the optimization in Sec.~\ref{sec:experiment}.

\subsection{Implementation Details}
In data preprocessing, $\dmargin$ is set to $0.15$ for scale normalization.
The contact and motion models share the same architecture hyperparameters, \ie with eight transformer encoder layers and a latent dimension of 512. 
Both models are trained with 50 diffusion steps on the ARCTIC~\cite{fan2023arctic} dataset for 200 epochs, using the Adam optimizer~\cite{diederik2014adam} of learning rate $1e^{-4}$ with a cosine learning rate scheduler~\cite{loshchilov2016sgdr}.
DDPM noise schedule~\cite{ho2020denoising} is adopted and the models directly predict clean samples. 
In addition, Exponential Moving Average (EMA) models~\cite{he2020momentum} are used for better stability. 
The motion model has a contact condition dropout rate of 0.5. 
For classifier-free guidance~\cite{ho2021classifierfree}, the guidance scale $\cfgscale$ is set to $0.5$.
We determine the contact map guidance scale $\cgscale$ by the gradient norm, \ie $\cgscale = \frac{1}{\left\Vert\nabla \hat{\handkpdirvec}_{(\diffusionindex)} \right\Vert}$.
Training is completed in less than two days on a single A40 GPU.
The post-processing is performed for 100 iterations, with $\ppweight_{\text{proj}}=100, \ppweight_{\text{pen}}=10$ and $\ppweight_{\text{acc}}=1000$ on ARCTIC. We set $\ppweight_{\text{acc}}$ to $10^4$ for HOI4D~\cite{Liu_2022_CVPR}. 
\section{Experiments}\label{sec:experiment}
% Double column figure.
 \begin{figure*}[!hbt]
 \centering
    %[trim={left bottom right top},clip]
    % ,trim=006mm 017mm 006mm 016mm,clip=true
	\includegraphics[width=1\linewidth]{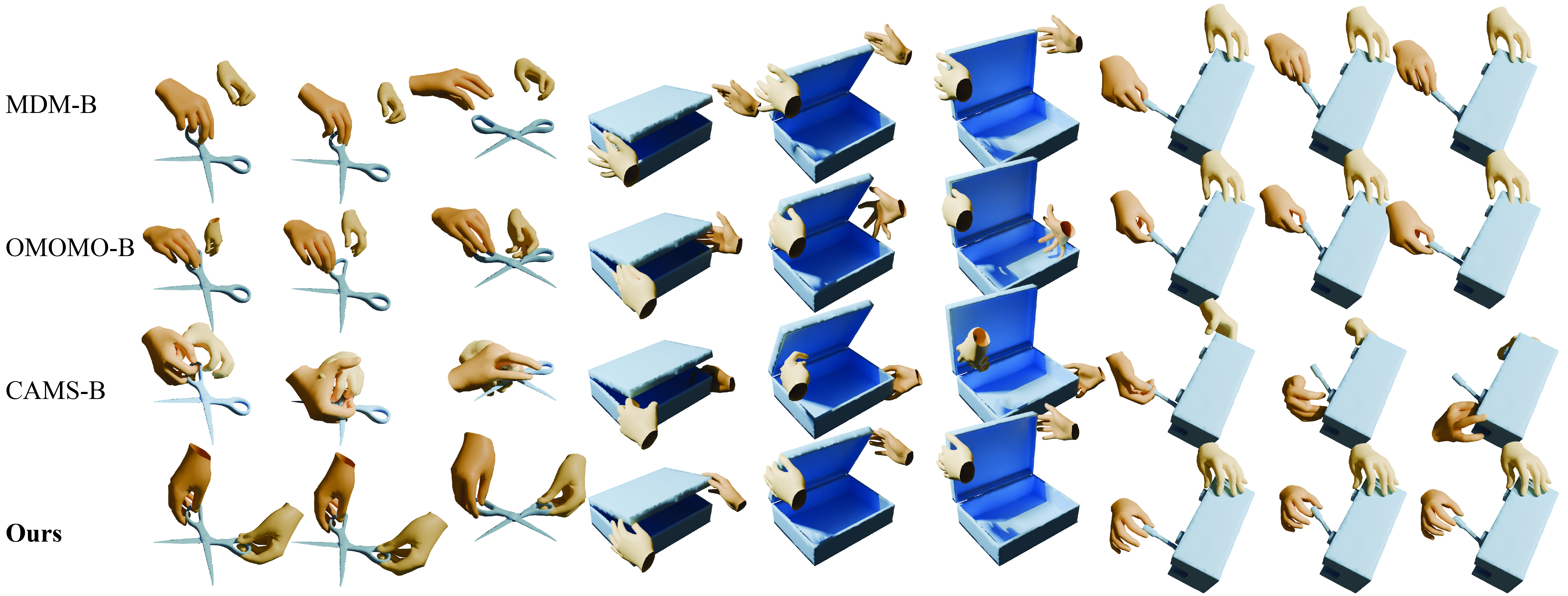}
    \vspace{-3mm}
        \caption
	{
        \textbf{Qualitative Comparison.} 
        \mdmb{} struggles with establishing accurate contact, as seen in the hand-object gap in the scissors and the box example.
        \omomob's rigid contact constraints make it prone to failure, especially with large wrist movements, like opening a box.
        \camsb{} failed to generate plausible motions, since its stage-wise contact targets under-constrain MANO fitting in dynamic settings with complex contact patterns and diverse object trajectories.
        }
	\label{fig:qualitative_comparison}
\end{figure*}
%
%%%%%%%%%%%%%%%%%%%%%%%%%%%%%%%%%%%%%
%
\begin{figure*}[!hbt]
    \centering
    \vspace{-4mm}
    \captionsetup{type=figure}
    \includegraphics[width=1\linewidth]{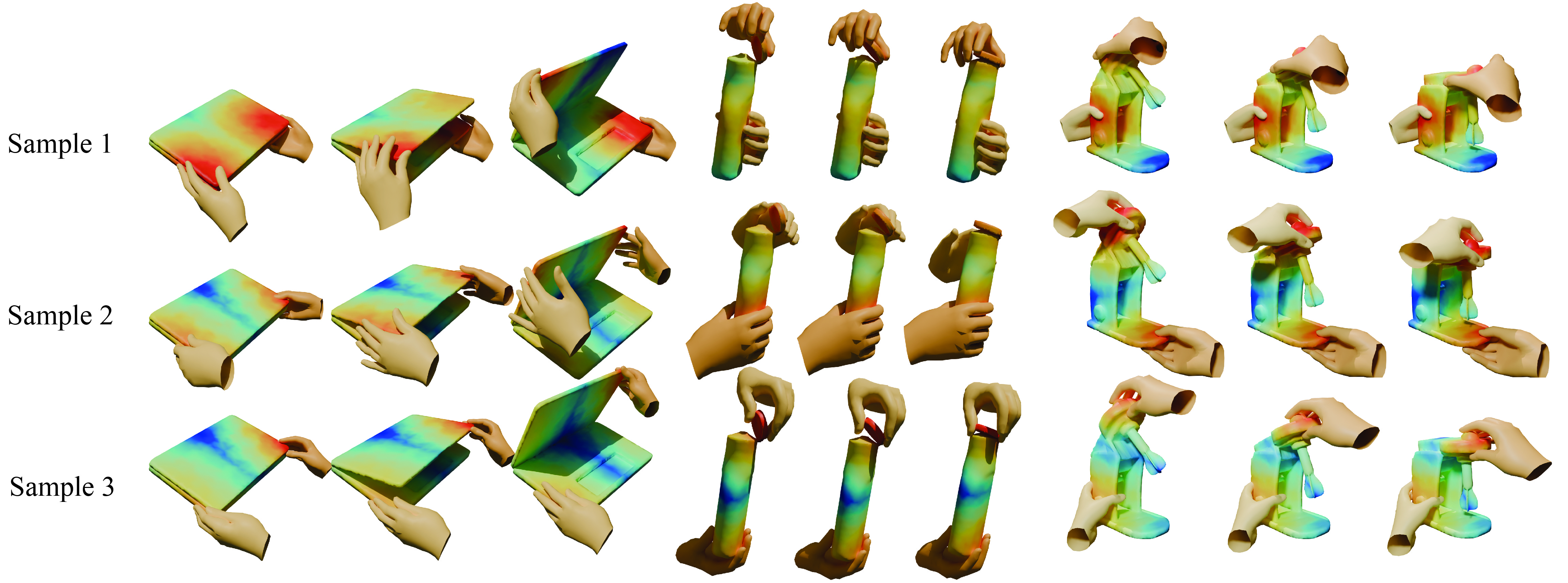}
    \vspace{-4mm}
    \captionof{figure}{ 
    \textbf{Diverse Results.}
     We show diverse bimanual sequences together with the predicted contact maps on the laptop, ketchup, and mixer given the same unseen trajectory per object.
     Our method generates accurate finger placements guided by the predicted contact maps. 
    }
    \label{fig:diverse_hand}
    \vspace{-6mm}
\end{figure*}
%
%%%%%%%%%%%%%%%%%%%%%%%%%%%%%%%%%%%%%
%
\paragraph{Datasets.}
We evaluate our method on the ARCTIC dataset ~\cite{fan2023arctic}, which contains fully annotated mesh sequences for bimanual interactions with 11 articulated objects. 
For each object, we use four motion sequences as the test set and the rest as training sequences. 
In total, we have 257 training sequences and 44 test sequences. 
In addition, we evaluate our method on HOI4D~\cite{Liu_2022_CVPR}, a large-scale dataset containing 3D annotations of articulated object movements and hand poses. 
We follow the evaluation protocol of Zheng \etal~\cite{Zheng_2023_CVPR_cams} and use two provided categories, pliers and scissors, with the same train and test split.

\vspace{-5mm}
\paragraph{Evaluation Metrics.}
Quantitatively measuring synthesized motion has been a challenging pursuit.
We evaluate the methods on various metrics, each targeting a specific aspect of motion generation.
The multi-modality metric, which measures the method's ability to generate diverse results for the same object trajectory, is computed using the mean average pairwise distance between all generated hand vertices by sampling 10 times for the same trajectory (denoted as ``Mul'' in \cref{tab:quantitative_baselines} and \cref{tab:ablation}).
To evaluate the geometric feasibility of the synthesized motions, we assess the extent of penetration and contact feasibility.
``Pen 1cm'' is the percentage of motion frames with hand vertex penetration, using a 1cm threshold.
``CM'' measures the $l1$ distance of the contact map derived from the
generated hand motions from the predicted contact map. This metric is only applicable to our ablations.
``Con'' measures the percentage of motion frames with object contact and ``Art'' measures the percentage of motion frames where the hand is in contact with the articulated part, out of the frames with object articulation changes. 
We also compute hand vertex penetration percentage, contact, and articulation consistency following \cams~\cite{Zheng_2023_CVPR_cams}'s protocol on HOI4D.
\vspace{-3mm}
\paragraph{Baselines.} 
Except for the concurrent work~\cite{zhang2024manidext}, no prior works have tackled bimanual motion synthesis for articulated objects given object trajectories under identical assumptions. 
Therefore, we propose the following modifications to various baselines: 
\begin{itemize}
    \item \cams~\cite{Zheng_2023_CVPR_cams} is a category-specific method that produces single-hand motions. \camsx{} denotes the cross-category model trained by us on HOI4D and \camsb{} is the bimanual model we adapted for ARCTIC.
    \item MDM~~\cite{tevet2023human} is a pioneer work for diffusion-based motion synthesis. We change text-based conditioning to object trajectory conditioning using our normalized part-based BPS features and denote this variant as \mdmb. We apply the same adaptation to the single-hand setting in the HOI4D dataset and refer to this variant as MDM-U.
    \item OMOMO~\cite{li2023object} is a whole-body method, which generates human-object interaction without finger articulations. In our adaptation \omomob, we generate hand joints in stage one with contact constraints applied to all the joints. In stage two, we predict the over-parameterized hand motions conditioned on joints. The single-hand variant for HOI4D dataset is denoted as MDM-U.
\end{itemize}
In addition, we follow \cams{} and include GraspTTA~\cite{jiang2021hand} and ManipNet~\cite{zhang2021manipnet} for comparisons on the HOI4D dataset. For more details, we refer to the \supmat.
% -----------
%%%%%%%%%%%%%%%%%%%%%%%%%%%%%%%%%%%%%%%%%%%%%%
%
\begin{table}[t]
\centering
\small
\resizebox{\linewidth}{!}
{
\begin{tabular}{lrrrrrrrrr}
    \toprule
    Method & Mul $(\text{cm})\uparrow$ & Accel $\left(\frac{\text{cm}}{\text{s}^2}\right) \downarrow$ & Pen 1cm (\%) $\downarrow$ & Con (\%) $\uparrow$ & Art (\%) $\uparrow$   \\
    \midrule
    GT &-&0.17848&1.0398&95.138&94.563\\ 
    \midrule
    \camsb  &\cellcolor{first}{8.5602}  &\cellcolor{first} 0.11959 &42.519 &\cellcolor{second}{98.915} &76.704\\
    \mdmb  &0.55459 &0.27666 &66.71 &93.657 &73.734
    \\
    \omomob &0.038338 &0.1969 &\cellcolor{second}{30.435} &96.917 &80.094
    \\
    \textbf{Ours}& \cellcolor{second}6.9093 &\cellcolor{second}0.18846 &\cellcolor{first}{2.0346} &\cellcolor{first}99.629 &\cellcolor{first}85.572\\
    \bottomrule
\end{tabular}
}
\caption{
\textbf{Quantitative Comparison on ARCTIC.} Our method outperforms the state of the art in penetration, contact, and articulation. Even though \camsb{} scores better in the multimodality and acceleration, it exhibits low interaction plausibility, as seen in high penetration percentage and qualitative results in \cref{fig:qualitative_comparison}.
} 
\vspace{-5mm}
\label{tab:quantitative_baselines}	
\end{table}

\begin{table}[hbt!]
\centering
\small
\resizebox{\linewidth}{!}
{
\begin{tabular}{lcccccccccc}
    \toprule[2pt]
    &&\multicolumn{3}{c}{\textbf{Pliers}} & \multicolumn{3}{c}{\textbf{Scissors}} \\ 
    \midrule
    && Pen $(\%)\downarrow$ & Con. Score $\uparrow$ & Art. Score $\uparrow$ & Pen $(\%)\downarrow$ & Con. Score $\uparrow$ & Art. Score $\uparrow$   \\
    \midrule
    &Ground Truth & 0.000 & 1.000 & 1.000 & 0.046 & 1.000 & 0.970 \\ 
    \midrule
    \multirow{8}{*}{Cat.Spec.} &&&&&\\
    &GraspTTA & 0.555 & 0.779 & 0.420 & 0.454 & 0.993 & 0.849 \\
    &GraspTTA w/ opt & 0.294 & 0.727 & 0.321 & 0.812 & 0.994 & 0.959 \\ 
    &ManipNet & 0.548 & 0.984 & 0.892 & 0.391 & 0.917 & 0.417 \\
    &ManipNet w/ opt & 0.387 & 0.890 & 0.738 & 0.131 & 0.831 & 0.333 \\
    &\cams{} w/ opt  & 0.563 & 0.916 & 0.393 & 0.590 & 0.997 & 0.850 \\
    &\cams & 0.004 & 1.000 & 1.000 & 0.080 & 0.999 & 0.989 \\ 
    \midrule
    \multirow{3}{*}{Unified}&&&&&\\
    &\camsx &\cellcolor{first}0.017 & 0.485&0.015 & \cellcolor{first}0.198&0.858 & 0.167 \\
    &MDM-U w/ opt & 0.225 & 0.767 & 0.090 &  \cellcolor{second}{0.224} & 0.994 & \cellcolor{first}0.999 \\
    &OMOMO-U w/ opt & 0.935 & 0.838 &\cellcolor{first}{0.829} &0.581 &0.990 &0.738 \\
    &Ours w/ opt & 0.464 & \cellcolor{second}0.870 & 0.595 & 1.204 & \cellcolor{first}{1.000} & \cellcolor{second}0.887 \\
    &\textbf{Ours} &\cellcolor{second}0.044 &\cellcolor{first}{0.966} &\cellcolor{second}{0.597} & 0.591& \cellcolor{first}{1.000}& 0.853  \\
\bottomrule
\end{tabular}
}
\caption{\textbf{Evaluation on the HOI4D Dataset.} 
We show comparisons in the category-specific setting (denoted as ``Cat.Spec'') and the cross-category setting where a unified model is trained (denoted as ``Unified'').
The numbers for ``Cat.Spec'' are taken from \cams~\cite{Zheng_2023_CVPR_cams}.
Our method outperforms \camsx{}, and performs comparatively with methods trained in a category-specific way.
}
\vspace{-6mm}
\label{tab:quant_hoi4d}
\end{table}

%%%%%%%%%%%%%%%%%%%%%%%%%%%%%%%%%%%%%%%%%%%%%%
%
\begin{table}[hbt!]
\centering
\small
\resizebox{\linewidth}{!}
{
\begin{tabular}{lrrrrrrrrr}
    \toprule
    &Method & Mul $(\text{cm})\uparrow$ & Accel $\left(\frac{\text{cm}}{\text{s}^2}\right) \downarrow$ & Pen 1cm (\%) $\downarrow$ & Con (\%) $\uparrow$ & Art (\%) $\uparrow$ & CM$(\text{cm}) \downarrow$  \\
    \midrule
    &GT &-&0.17848&1.0398&95.138&94.563&-\\ 
    \midrule
    \multirow{5}{*}{Rep.} &&&&&&\\
    &U-BPS &6.2551 &0.28275 &\cellcolor{second}17.542 &97.214 &\cellcolor{first}85.642 &1.1072  \\
   & NPA-BPS &6.317 &0.28537  &\cellcolor{first}17.508 &97.442 &82.04 &\cellcolor{second}1.0789  \\
   & MANO-Rep &\cellcolor{second}6.4781 &\cellcolor{first}0.27243 &22.201 &95.255 &76.814 &1.6379 \\
   & NP-BPS w/o $\globalstateseq$ &6.4149 &0\cellcolor{second}.28233 &20.414 &\cellcolor{second}98.093 &82.079 &\cellcolor{first}1.0562  \\
   & \textbf{NP-BPS} &\cellcolor{first}6.97928&0.31398 &20.273 &\cellcolor{first}98.481 &\cellcolor{second}83.089 &1.1505 \\ %
    \midrule
    \multirow{4}{*}{Contact.} &&&&&&\\
   & w/o $\contactseq$ &4.4894&0.31034 &\cellcolor{second}9.4481 &96.129 &79.591 &-\\ %
   & w $\contactseq$&\cellcolor{first}6.97928&0.31398 &20.273 &\cellcolor{second}98.481 &83.089 &\cellcolor{second}1.1505 \\ %
   & w $\contactseq$ + CG & \cellcolor{second}6.9551 &\cellcolor{second}0.30371  &16.496 &97.351 &\cellcolor{second}84.227 &\cellcolor{first}1.1284  \\
  &  \textbf{w $\contactseq$+CG+Opt}& 6.9093 &\cellcolor{first}0.18846 &\cellcolor{first}{2.0346} &\cellcolor{first}99.629 &\cellcolor{first}85.572 &1.1778 \\
    \bottomrule
\end{tabular}
}
\caption{Ablations for various object and hand representations and ways to utilize contact information based on the ARCTIC dataset. 
        The experiment in bold is our proposed design. 
        } 
        \vspace{-6mm}
\label{tab:ablation}	
\end{table}

\subsection{Quantitative Results}\label{sec:experiment-quant-result}
We tabulate the quantitative comparison of the methods in ~\cref{tab:quantitative_baselines} and ~\cref{tab:quant_hoi4d} for the ARCTIC and HOI4D datasets, respectively. 
Our method outperforms \mdmb{} and \omomob{} in all metrics. 
Even though \camsb{} scores better in multi-modality, acceleration, and contact, we show qualitatively that \camsb{} struggles to produce natural and plausible motions in \cref{sec:experiment-qualitative-result}.
\par
In the single-hand setting on HOI4D (\cref{tab:quant_hoi4d}), our cross-category model performs comparatively with the category-specific baselines.
In the cross-category setting, we outperform \camsx{} in terms of articulation and contact consistency by a large margin, highlighting the advantage of our method to handle a variety of geometries in a unified manner.
We refer the reader to the \supvid for a holistic assessment of our results.
\subsection{Perceptual User Study}\label{sec:experiment-user-study-result} 
\begin{figure}[!t]
	\includegraphics[width=\linewidth]{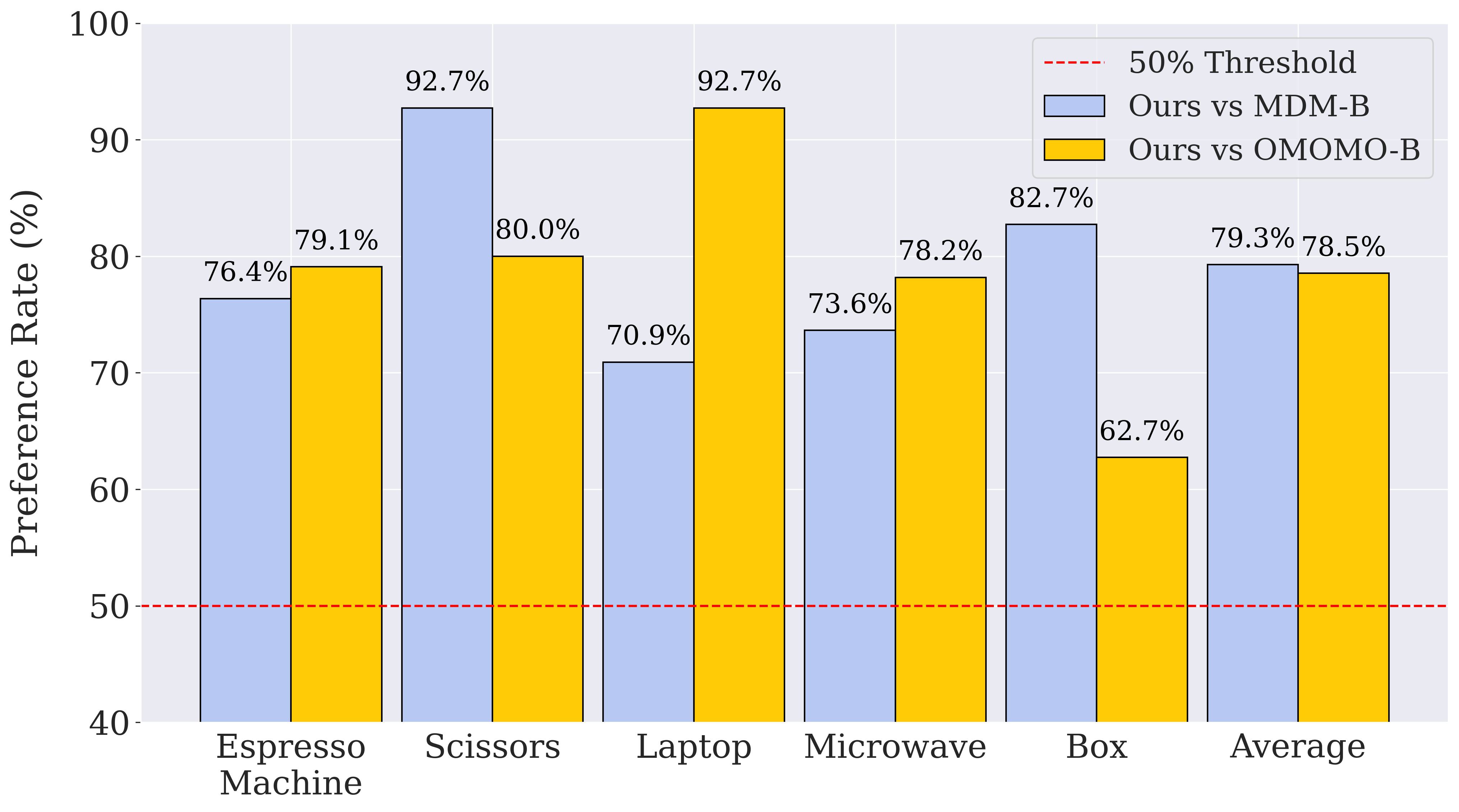}
	\caption{
        \textbf{User Study Results}. We show the preference rate of \methodname{} against \mdmb{} and \omomob. Our method outperforms the existing state of the art for all objects covered in the user study.  }
            \vspace{-5mm}
	\label{fig:user_study}
\end{figure}

The interaction plausibility is difficult to assess using quantitative metrics alone. 
Hence, we conducted a perceptual user study with 55 human respondents to evaluate our generated motions compared with \omomob{} and \mdmb. 
We exclude \camsb{} for the user study, as its motion quality is significantly subpar evident in \cref{fig:qualitative_comparison} and our \supvid. 
The user study contains 40 pairs of animations, covering five objects and four object trajectories randomly sampled from the test set.
We split the user study into two subgroups, each covering two out of the four object trajectories with 20 questions. 
In each question, we present the participants with two animations with the same object trajectory, one of which is generated by our method.
The survey has a force-choice style with the following question: \textit{Which animation has a more natural hand motion that aligns better with the object trajectory?}
The animations are interactive, allowing the user to zoom or rotate the view to access the quality accurately.
We calculate p-values (z-test) for our comparisons and observe statistically significant results with $p<10^{-3}$ for all baselines. 
We demonstrate that the users prefer our motions for every object in ~\cref{fig:user_study}. 

\subsection{Qualitative Results}\label{sec:experiment-qualitative-result}
We show qualitative comparisons in ~\cref{fig:qualitative_comparison}
and diverse samples from our method in ~\cref{fig:diverse_hand}.
\methodname{} generates more natural and physically realistic motions for small objects that require precise contact region geometry understanding, like the ``grabbing the scissors'' example, and large objects with significant articulation movements, like the ``opening box'' example in \cref{fig:qualitative_comparison}.
~\cref{fig:diverse_hand} shows three diverse samples of hand motions given the same object trajectory. 
\subsection{Ablations} \label{sec:ablations}
We present ablation studies to investigate the effect of our design choices and report the results
in \cref{tab:ablation}.
%. 
We split the ablations into two sections, with the ``Rep'' section ablating the different object representations and hand representations.
The BPS representations in \cref{tab:ablation} include Unnormalized BPS (U-BPS), normalized part-agnostic BPS (NPA-BPS), and our proposed normalized part-based BPS (NP-BPS), all trained with contact conditions. 
We also ablate the effect of removing global states $\globalstateseq$ from (NP-BPS w/o $\globalstateseq$), leading to the lack of the object global movement awareness in the contact generation and motion generation model.
As an alternative hand representation, MANO-Rep refers to MANO 6D pose parameters with joint positions.
Notably, we do not apply contact map guidance and post-refinement in this set of experiments to isolate the effect caused by object and hand representations.
The ``Contact'' section demonstrates the effect of the contact condition, (\ie w/o $\contactseq$ versus w $\contactseq$), contact map guidance (w $\contactseq$ + CG), and the optimization-based refinement (w $\contactseq$ + CG + Opt). 
\par
The following observations can be drawn from the \cref{tab:ablation}.
NP-BPS leads to the highest multi-modality and contact percentage compared with the alternative BPS sampling strategies.
MANO-Rep leads to worse penetration, contact, and articulation percentages, highlighting its limitations compared to our proposed hand representation.
NP-BPS w/o G performs better in acceleration and contact map discrepancy by focusing on articulation-aware hand motions, while excluding global states, which we show leads to physically implausible motions in the \supvid.
In contact ablations, contact conditioning leads to a higher multi-modality, contact, and articulation percentage.
Contact guidance helps the hand motions better align with the contact maps, evidenced by a lower contact map discrepancy. 
Our optimization-based refinement significantly reduces the penetration percentage and the hand motion acceleration while maintaining contact with the objects.
For qualitative ablations, please refer to the video.

\section{Conclusion}\label{sec: conclusion}
\par \noindent \textbf{Limitations.} Although fairly robust to novel object trajectories, our method is restricted to the limited number of object categories as provided in the datasets (ARCTIC and HOI4D). 
In the real world, however, one would like to generalize to new (and open-vocabulary) objects in a zero-shot manner.
We believe leveraging the common-sense knowledge of existing multi-modal large language models would facilitate such generalization~\cite{li2024genzi, hong20233d}.
Our method could also benefit from the incorporation of faster diffusion sampling approaches such as DDIM~\cite{song2021denoising}, or Latent Diffusion Modeling~\cite{chen2023executing} to facilitate adoption in artistic creation processes with limited time budgets.
\par 
This paper introduced \methodname{}, a new bimanual motion synthesis method assuming a trajectory of an articulated 3D object as input.
Our proposed feature representation leads to high diversity in generated motions, providing the flexibility for 3D artists and animators to sample multiple plausible interactions for a single object trajectory.
In both quantitative metrics and the user study, our approach outperforms competing methods in terms of naturalness and physical plausibility, paving the way for more realistic and user-friendly hand-object animation.

\newpage
\par \noindent \textbf{Acknowledgements.} This work was supported by the Saarbrücken Research Center for Visual Computing, Interaction and Artificial Intelligence (VIA) and the Deutsche Forschungsgemeinschaft (DFG, German Research Foundation) – GRK 2853/1 “Neuroexplicit Models of Language, Vision, and Action”. The authors would like to thank Krzysztof Wolski for the help on Blender visualizations and Hui Zhang for the helpful discussion on setting up the ArtiGrasp baseline.
    \small
    \bibliographystyle{ieeenat_fullname}
    \bibliography{biblio}
% \clearpage
% {

% }

\normalsize
\clearpage
\appendix
\maketitlesupplementary
\setcounter{figure}{0}
\setcounter{table}{0}

\renewcommand\thetable{\Roman{table}}
\renewcommand\thefigure{\Roman{figure}}
In this document, we first present a conceptual comparison of our setting and approach in \cref{supp: concept}. We provide additional details such as data processing (\cref{supp: data_process}), baseline adaptation (\cref{supp: baseline_adaptation}), and additional results (\cref{supp: results}). Please refer to the \supvid for animations.
\section{Conceptual Comparison}\label{supp: concept}
Our method relies on fewer assumptions than the prior works, as shown in ~\cref{tab:concecptual_comparison}.

\section{Data Processing} \label{supp: data_process}
We follow the convention of the ARCTIC dataset~\cite{fan2023arctic}, defining the canonical space as the configuration where the articulation axis aligns with the negative z-axis. For scale normalization, we apply a heuristic to determine an articulation angle that positions the object at a state likely to maximize its distance from the origin. Specifically, we set the articulation angle to $\frac{\pi}{2}$.
For the mixer and capsule machine, and to 0 for the scissors and espresso machine. For all other objects, we set the articulation angle to $\pi$.

\section{Baseline Details} \label{supp: baseline_adaptation}
\paragraph{OMOMO Adaptation.} 
In the original full-body setting, OMOMO~\cite{li2023object} predicts only the wrist positions in stage one.
Since the OMOMO dataset lacks finger articulation data, the wrist, being the closest joint to the object, is the natural choice for applying contact constraints. In contrast, in our hand-only setting, all joints have the potential to interact with objects. Limiting contact constraints to the wrist in this context would be suboptimal. Therefore, we design stage one to predict all hand joints, applying contact constraints to each joint. In stage two, we refine the motion predictions by estimating the hand poses, conditioned on all joints.

\paragraph{ArtiGrasp.} 
We also re-trained ArtiGrasp~\cite{zhang2024artigrasp} on our train/test split and evaluated the dynamic object grasping and articulation task which performs grasping and articulation in separate stages. 
Since the object's initial state has to be supported by the table in the simulator, we set the relative change of the object state to be the same without violating the physical constraint (eg. the goal state should not penetrate the table). 
ArtiGrasp cannot reach the object goal state reliably at every run, unavoidably, the actual object trajectory from the physics simulator will deviate significantly from ours.
Moreover, ArtiGrasp employs heuristics transitioning from grasping to articulation, such as dropping the object on the table and moving the hands apart before articulating, resulting in low contact and articulation percentage.  
Due to the difficulty in standardizing the setting, we exclude ArtiGrasp from our quantitative and qualitative comparisons.

\section{Additional Results} \label{supp: results}
Besides providing the penetration percentage at 1cm threshold in the main paper, we additionally provide it at 5mm as shown in \cref{tab:quantitative_addition}.
%%%%%%%%%%%%%%%%%%%%%%%%%
\begin{table} 
    \small 

	\centering
	\begin{tabular}{lcccccc}
        \toprule
        Method & \makecell{Articulated \\Objects} & \makecell{Bimanual} & \makecell{No Grasp\\Ref.} & Unified \\
        \midrule
        ManipNet~\cite{zhang2021manipnet}      & \xmark  & \cmark    & \cmark    & \cmark         \\
        GOAL~\cite{taheri2021goal}             & \xmark  & \cmark    & \cmark    & \cmark         \\
        IMOS~\cite{ghosh2022imos}              & \xmark  & \cmark    & \xmark    & \cmark         \\
        MACS~\cite{MACS2024}                   & \xmark  & \cmark    & \cmark    & (\xmark)          \\
        D-Grasp~\cite{christen2022dgrasp}      & \xmark  & \xmark    & \xmark    & \cmark       \\
        ArtiGrasp~\cite{zhang2024artigrasp}    & \cmark  & \cmark    & \xmark    & \cmark        \\	
        CAMS~\cite{Zheng_2023_CVPR_cams}	    & \cmark  & \xmark    & \xmark    & \xmark        \\
	\textbf{\methodname}  	                & \cmark  & \cmark    & \cmark    & \cmark     \\
        \bottomrule
	\end{tabular}
        \caption{\textbf{Conceptual Comparison to Prior Works.}
        We highlight that our work is the only one, which provides all desired functionalities.
        \textbf{No Grasp Ref.} means that neither initial pose nor goal pose are given as input.
        \textbf{Unified} refers to a single model that can handle various object categories. 
        MACS is only trained on spheres, hence a bracket is added for the checkmark under Unified.
        }
    \label{tab:concecptual_comparison}
\end{table}
%%%%%%%%%%%%%%%%%%%%%%%%%

\begin{table}[hbt]
\centering
\small
\begin{tabular}{lr}
    \toprule
    Method  & Pen 5mm (\%) $\downarrow$ \\
    \midrule
    GT & 30.4\\ 
    \midrule
    \camsb & 87.5 \\
    \mdmb & \cellcolor{second} 66.7\\
    \omomob & 74.9 \\
    \midrule
    \textbf{Ours} & \cellcolor{first}32.8\\
    \bottomrule
\end{tabular}
\caption{Penetration percentage at the 5mm threshold} 
\label{tab:quantitative_addition}	
\end{table}

\begin{table*}[h!]
\centering
\small
\resizebox{\linewidth}{!}
{
\begin{tabular}{lrrrrrrrrrrrr}
\toprule
& Average & Microwave & Phone & Box & Ketchup& Mixer & Waffle Iron & Capsule Machine & Notebook & Scissors & Laptop& Espresso Machine \\ \hline
\hline
U-BPS-Top & 0.546 & 0.608 & 0.244 & 0.454 & 0.387 & 0.838 & 0.705 & 0.513 & 0.589 & 0.401 & 0.484 & 0.746 \\ 
PA-BPS-Top & \cellcolor{second}0.342 & \cellcolor{second}0.507 & \cellcolor{first}0.137 & \cellcolor{second}0.415 & \cellcolor{second}0.221 & \cellcolor{second}0.377 & \cellcolor{second}0.427 & \cellcolor{second}0.394 & \cellcolor{second}0.288 & \cellcolor{second}0.093 & \cellcolor{first}0.341 & \cellcolor{second}0.533 \\ 
\textbf{P-BPS-Top} & \cellcolor{first}0.258 & \cellcolor{first}0.327 & \cellcolor{second}0.152 & \cellcolor{first}0.373 & \cellcolor{first}0.114 & \cellcolor{first}0.336 & \cellcolor{first}0.413 & \cellcolor{first}0.185 & \cellcolor{first}0.265 & \cellcolor{first}0.081 & \cellcolor{second}0.349 & \cellcolor{first}0.216 \\ 
\midrule
U-BPS-Bottom & 0.552 & 0.651 & 0.25 & \cellcolor{second}0.5 & 0.387 & 0.725 & 0.57 & 0.572 & 0.543 & 0.543 & 0.523 & 0.809 \\ 
PA-BPS-Bottom & \cellcolor{first}0.341 & \cellcolor{first}0.482 & \cellcolor{first}0.466 & \cellcolor{first}0.194 & \cellcolor{second}0.377 & \cellcolor{first}0.349 & \cellcolor{first}0.103 & \cellcolor{first}0.36 & \cellcolor{second}0.378 & \cellcolor{second}0.374 & \cellcolor{first}0.263 & \cellcolor{first}0.145 \\ 
\textbf{P-BPS-Bottom} & \cellcolor{second}0.38 & \cellcolor{second}0.645 & \cellcolor{second}0.173 & 0.507 & \cellcolor{first}0.232 & \cellcolor{second}0.46 & \cellcolor{second}0.368 & \cellcolor{second}0.472 & \cellcolor{first}0.27 & \cellcolor{first}0.094 & \cellcolor{second}0.395 & \cellcolor{second}0.536 \\ 
\midrule
U-BPS & 0.554 & 0.645 & 0.247 & 0.48 & 0.387 & 0.763 & 0.643 & 0.568 & 0.562 & 0.468 & 0.504 & 0.807 \\ 
PA-BPS & \cellcolor{first}0.32 & \cellcolor{first}0.487 & \cellcolor{first}0.14 & \cellcolor{first}0.444 & \cellcolor{first}0.199 & \cellcolor{first}0.366 & \cellcolor{second}0.404 & \cellcolor{first}0.35 & \cellcolor{second}0.272 & \cellcolor{first}0.099 &\cellcolor{first} 0.36 & \cellcolor{first}0.378 \\ 
\textbf{P-BPS} & \cellcolor{second}0.361 & \cellcolor{second}0.603 & \cellcolor{second}0.163 & \cellcolor{second}0.449 & \cellcolor{second}0.208 & \cellcolor{second}0.418 & \cellcolor{first}0.393 & \cellcolor{second}0.453 & \cellcolor{first}0.268 & \cellcolor{second}0.087 & \cellcolor{second}0.373 & \cellcolor{second}0.527 \\ 
\bottomrule
\end{tabular}
}
\caption{\textbf{Contact Map Error (in cm) due to BPS mapping.} We present the average and per-category contact map errors resulting from the sparse mapping of BPS features. Both part-agnostic BPS (PA-BPS) and the proposed part BPS (P-BPS) achieve a denser mapping compared to BPS features without scale normalization (U-BPS), resulting in smaller contact map errors. The proposed part-based BPS method further enhances mapping density for the top part of the object (which corresponds to the movable part in canonical space), by allocating equal feature dimensions to individual parts irrespective of their surface area.}
\label{tab:bps_contact_error}	
\end{table*}

\begin{figure*}[h]
    \centering
    \includegraphics[width=0.33\textwidth]{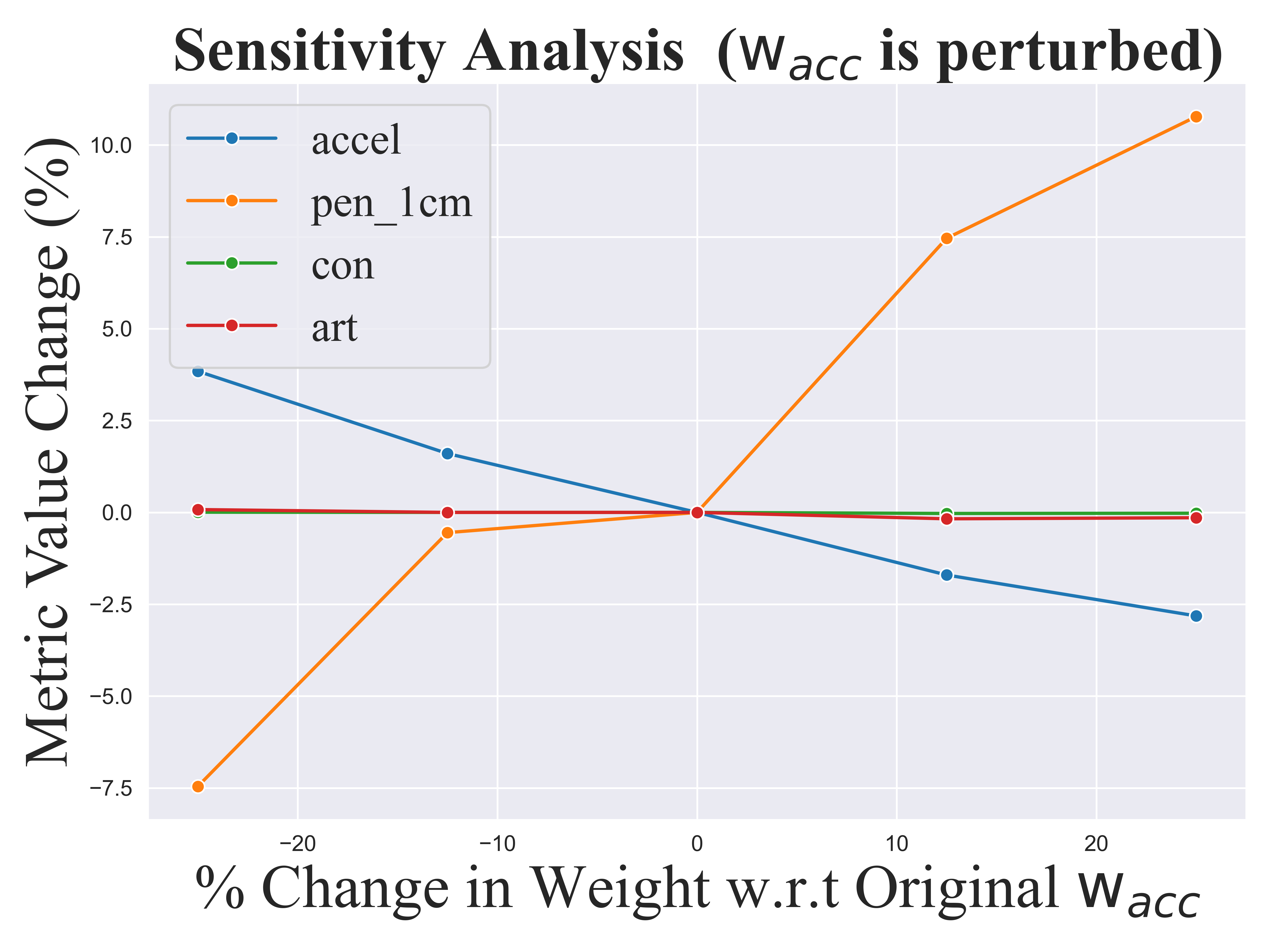}
    \includegraphics[width=0.33\textwidth]{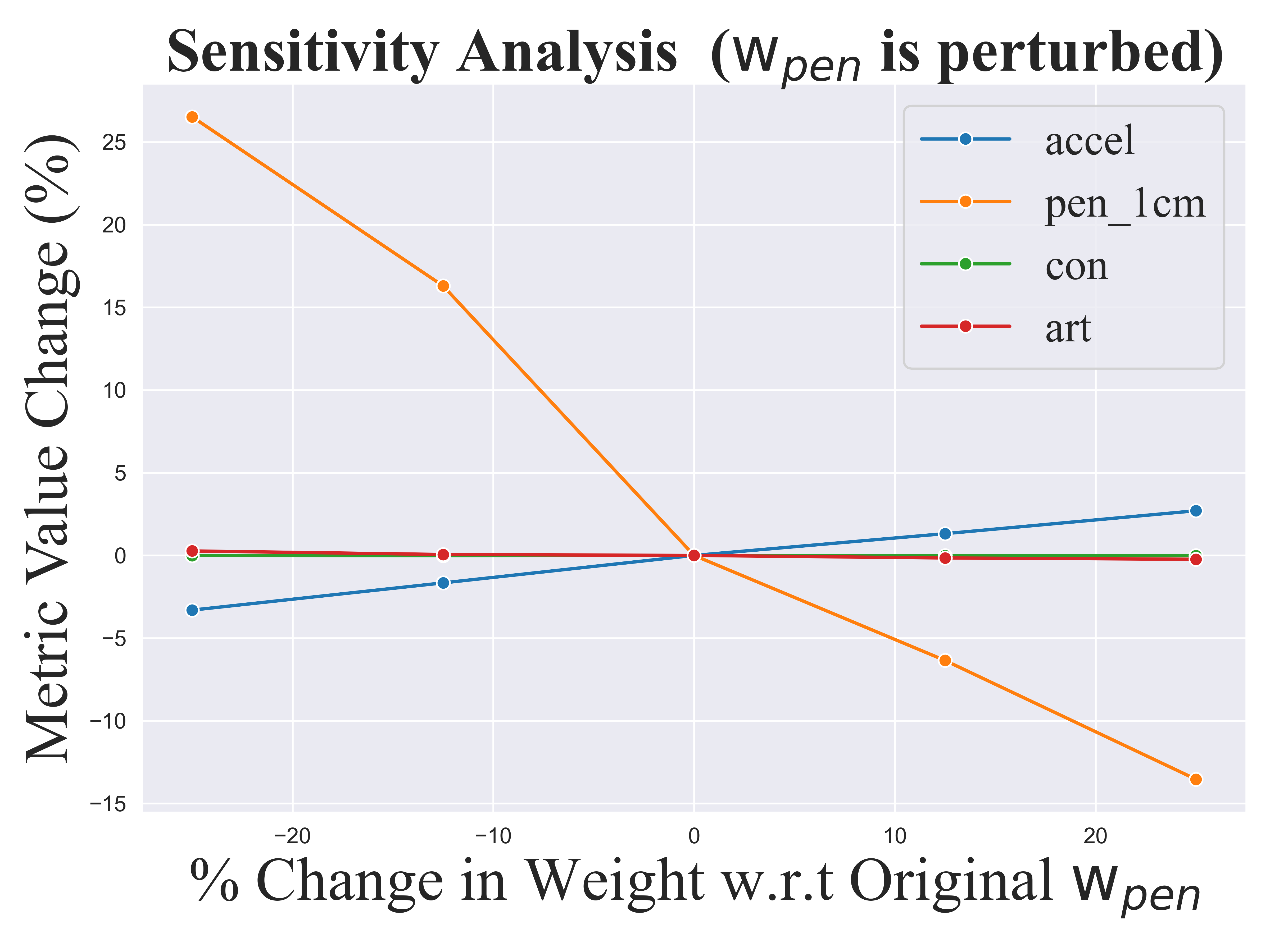}
    \includegraphics[width=0.33\textwidth]{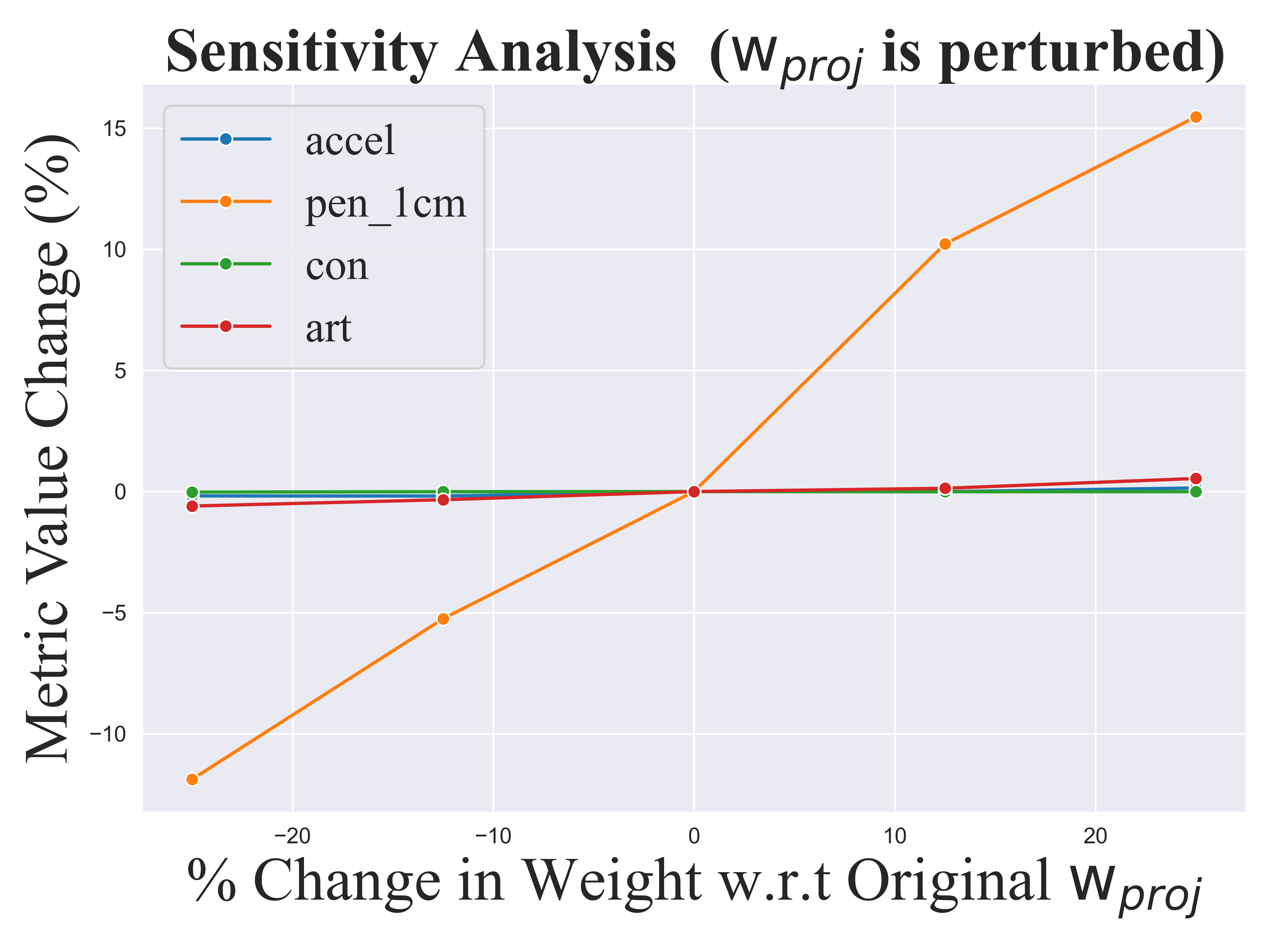}
    \caption{Sensitivity analysis for $w_{acc}$ (left plot), $w_{pen}$ (middle plot) and $w_{proj}$ (right plot). We perturb each hyperparameter within $\pm 25\%$ and report the changes in the acceleration, penetration, contact, and articulation metrics. }
    \label{fig:sensitivity}
\end{figure*}

\begin{figure*}
\centering
	\includegraphics[width=0.8\linewidth]{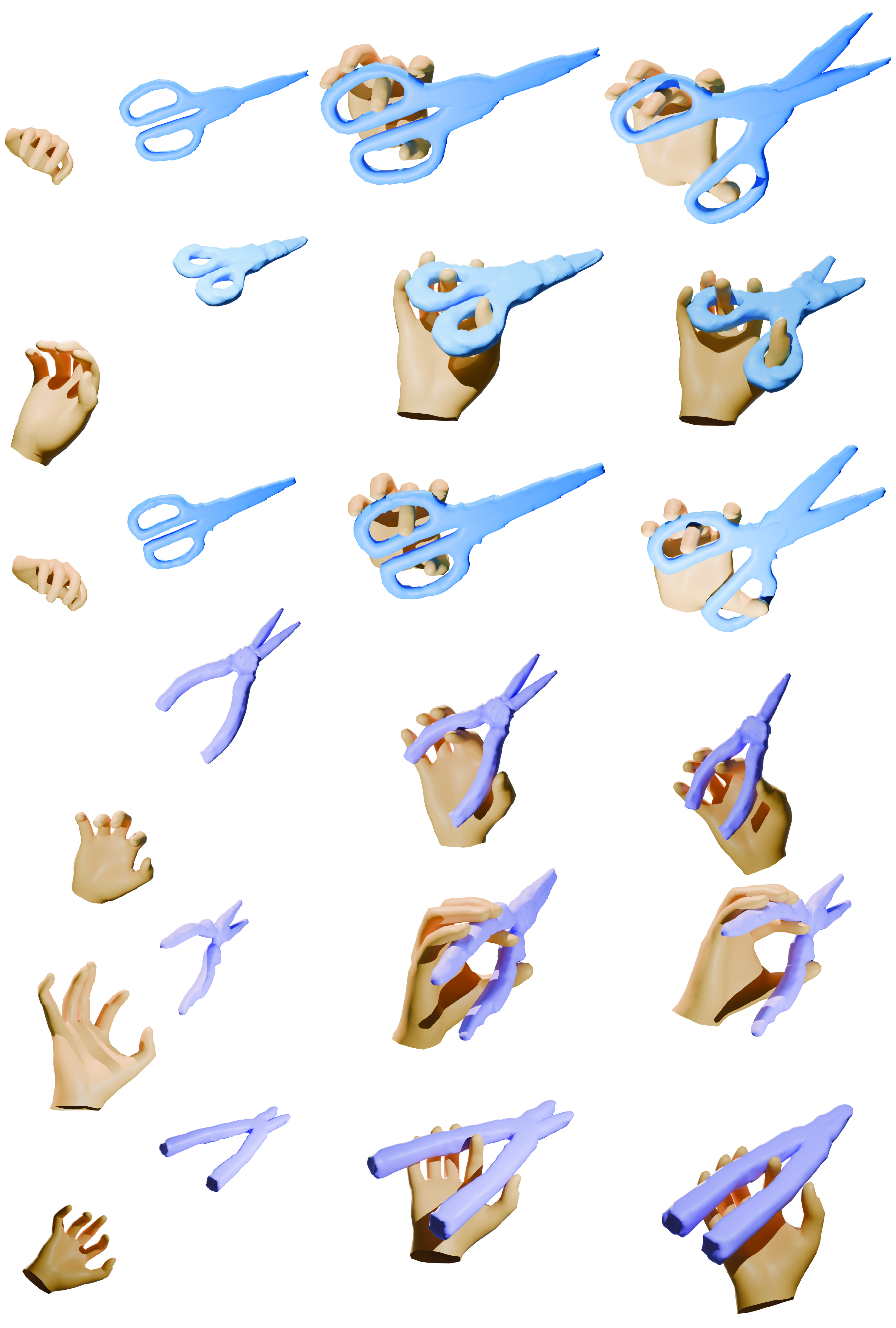}
	\caption{
        \textbf{Qualitative Results on HOI4D}. We present visualizations of results for six unseen objects from the HOI4D dataset. Each row illustrates three frames corresponding to the actions of approaching, lifting, and articulating. Notably, our model is trained in a cross-category way. }
	\label{fig:qualitative_hoi4d}
\end{figure*}
To show that we are not overfitting to the ground truth, we compute the five nearest neighbors in the training set for each test sequence based on object motions, with the first frame of object vertices centered at zero. We obtain a 15.08 cm average hand vertex distance with a 4.40cm average object vertex distance, showing that our generated motions differ from the training ground truth. Please see the \supvid for qualitative results.

In \cref{fig:sensitivity}, we show sensitivity analysis plots for and $w_{\text{acc}}$, $w_{\text{proj}}$ and $w_{\text{pen}}$ respectively by perturbing each hyperparameter by $\pm25\%$ of its original weight. We show the percentage change in the acceleration, articulation, contact, and penetration metrics for each plot. We observe that contact and articulation are not very sensitive to the hyperparameter perturbations, and there exists a trade-off between assigning a higher weight for $w_{\text{acc}}$ and assigning a higher weight for $w_{\text{pen}}$ as evident in the first 2 plots in \cref{fig:sensitivity}. A higher weight for $w_{\text{acc}}$ leads to better motion smoothness but it increases penetration, and vice versa, when we increase $w_{\text{pen}}$, the motion gets more jittery.  

%
%%%%%%%%%%%%%%%%%%%%%%%%%%%%%%%%%%%%%
%
\begin{figure*}[!hbt]
    \centering
    \captionsetup{type=figure}
    \includegraphics[width=1.0\linewidth]{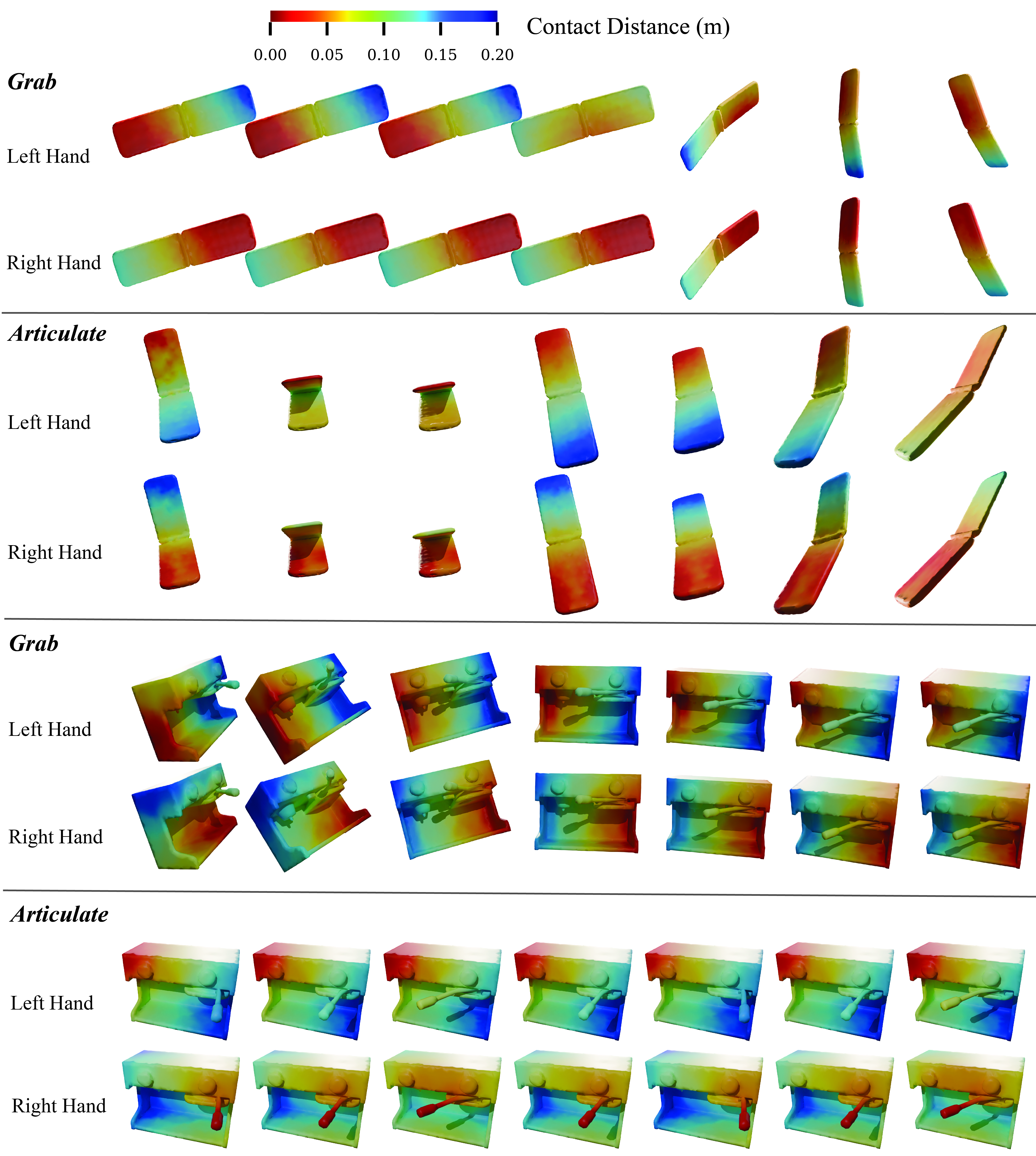}
    \captionof{figure}{
     \textbf{Contact map visualizations.} 
         We present visualizations of the predicted left and right contact maps for seven frames in a sequence. For each object, we include two examples: a  ``grab''  scenario, where the object's articulation remains unchanged, and an ``articulate'' scenario, where the object undergoes articulation.
        In the ``articulate'' examples, the contact region is established at the moving part and remains consistent throughout the articulation process. In contrast, the ``grab'' examples reveal shifts in the grasping patterns, suggesting that one hand holds the object while the other adjusts its contact point.
The Vector Heat method \cite{sharp2019vector} is employed to interpolate the contact values from the sampled object vertices to the full object surface. The predicted contact values are then normalized to a range between $0$ and $0.2$ meters. In the resulting visualization, red indicates that the hand should be close to the object's surface, while blue signifies that the hand is farther away.
     }
    \label{fig:contact_map}
\end{figure*}
%
%%%%%%%%%%%%%%%%%%%%%%%%%%%%%%%%%%%%%
Qualitatively, we visualize diverse contact maps our method generates in ~\cref{fig:contact_map}.
\cref{fig:qualitative_hoi4d} shows the generalization ability of our method to intra-class variations in the HOI4D dataset~\cite{Liu_2022_CVPR}. Our model is trained in a cross-category manner and we show the qualitative results for all six unseen objects. 
\section{BPS Analysis} \label{supp: bps_contact_error}
We present additional BPS feature analysis in \cref{tab:bps_contact_error}, by interpolating the contact values associated with sparse object vertices mapped by the basis points using \cite{sharp2019vector} and compute the L1 loss for the densified per vertex contact maps and the ground truth contact maps. A lower error reflects a denser BPS mapping and better geometric representation. The results are broken down into cross-category averages and object-specific errors, with errors reported for the top part, bottom part, and whole object. 
Both part-agnostic BPS (PA-BPS) and the proposed part-based BPS (P-BPS) achieve lower contact errors compared to unnormalized BPS (U-BPS) with the same BPS feature dimensions. 
PA-BPS achieves a lower average contact map error for the object's bottom parts as they tend to have a larger surface area in the ARCTIC dataset ~\cite{fan2023arctic}.
Notably, P-BPS reduces the contact map error for the objects' top parts (the movable component in our canonical space) by allocating equal feature dimensions to the top and bottom parts.

\end{document}